\documentclass{article}

\usepackage[preprint]{neurips_data_2024}

\usepackage[bottom]{footmisc}
\usepackage[utf8]{inputenc} 
\usepackage[T1]{fontenc}    
\usepackage{hyperref}       
\usepackage{url}            
\usepackage{booktabs}       
\usepackage{amsfonts}       
\usepackage{nicefrac}       
\usepackage{microtype}      

\usepackage{amssymb}            
\usepackage{mathtools}          
\usepackage{mathrsfs}           
\usepackage{graphicx}           
\usepackage{subcaption}         
\usepackage[space]{grffile}     
\usepackage{url}                

\usepackage{times}
\usepackage{tabularx, makecell}
\usepackage[utf8]{inputenc} 
\usepackage[T1]{fontenc}    
\usepackage{hyperref}       
\usepackage{url}            
\usepackage{booktabs}       
\usepackage{amsfonts}       
\usepackage{nicefrac}       
\usepackage{microtype}      
\usepackage{xcolor}     
\usepackage{natbib}

\usepackage{placeins}
\usepackage{enumitem}
\usepackage{comment}
\usepackage{todonotes}
\usepackage{blindtext}
\usepackage{caption}

\usepackage{amssymb}
\usepackage{amsbsy}
\usepackage{upgreek}
\usepackage{amsmath}
\usepackage{wrapfig}

\usepackage[frozencache, cachedir=.]{minted}

\newcommand{\eg}{\emph{e.g.}~} 

\newcommand{\ie}{\emph{i.e.}~} 

\newcommand{\cf}{\emph{cf.}~}

\newcommand{\ramcell}[1]{\fcolorbox{gray}{white}{$#1$}}

\newcommand{\smallpm}[1]{\mathbin{\scriptstyle\pm\scriptstyle#1}}

\title{HackAtari: Atari Learning Environments \\for Robust and Continual Reinforcement Learning}

\author{Quentin Delfosse$^{*,1}$\!\!\! \quad Jannis Blüml$^{*,1,2}$\!\!\! \quad  Bjarne Gregori$^1$\!\!\! \quad Kristian Kersting$^{1,2,3,4}$ \\ \\
$^1$AI and ML Group, Technical University of Darmstadt, Germany \\
$^2$Hessian Center for Artificial Intelligence (hessian.AI) \\
$^3$Centre for Cognitive Science of Darmstadt \\
$^4$German Research Center for Artificial Intelligence (DFKI) \\
\href{mailto:quentin.delfosse@cs.tu-darmstadt.de}{\texttt{quentin.delfosse@cs.tu-darmstadt.de}}, \href{mailto:blueml@cs.tu-darmstadt.de}{\texttt{blueml@cs.tu-darmstadt.de}}
}

\begin{document}

\maketitle

\begin{abstract}
Artificial agents' adaptability to novelty and alignment with intended behavior is crucial for their effective deployment. Reinforcement learning (RL) leverages novelty as a means of exploration, yet agents often struggle to handle novel situations, hindering generalization. 
To address these issues, we propose HackAtari, a framework introducing controlled novelty to the most common RL benchmark, the Atari Learning Environment. 
HackAtari allows us to create novel game scenarios (including simplification for curriculum learning), to swap the game elements' colors, as well as to introduce different reward signals for the agent. 
We demonstrate that current agents trained on the original environments include robustness failures, and evaluate HackAtari's efficacy in enhancing RL agents' robustness and aligning behavior through experiments using C51 and PPO. 
Overall, HackAtari can be used to improve the robustness of current and future RL algorithms, allowing Neuro-Symbolic RL, curriculum RL, causal RL, as well as LLM-driven RL. 
Our work underscores the significance of developing interpretable in RL agents. \footnote{Code available at \href{https://github.com/k4ntz/HackAtari}{https://github.com/k4ntz/HackAtari}, $^{*}$ Equal Contribution}
    
\end{abstract}

\section{Introduction}
\begin{figure}[t]
    \centering
    \includegraphics[width=1.0\linewidth]{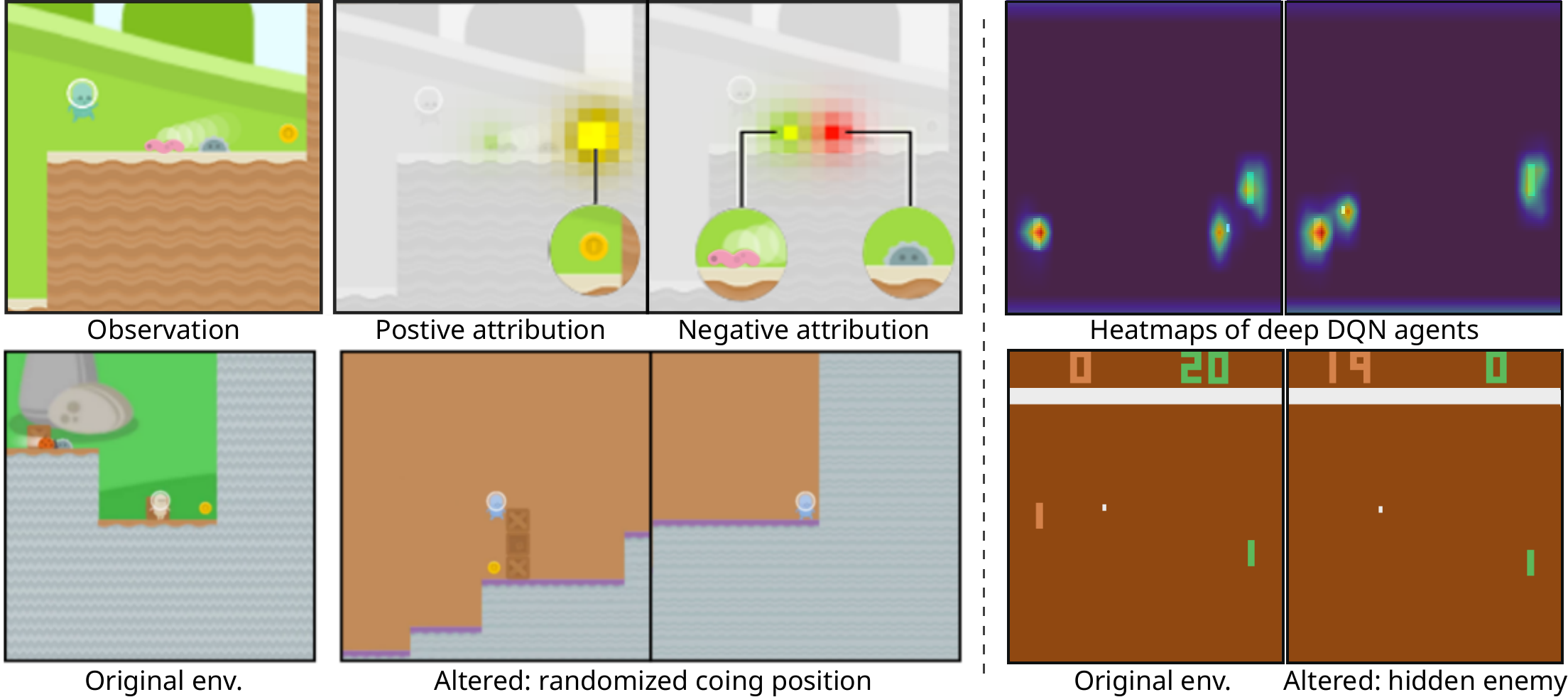}
    \caption{\textbf{Examples of misaligned agents.} 
    In Coinrun (left), agents learn to reach the end of the level, instead of the coin.
    In Pong (right), agents learn to follow the enemy instead of the ball.
    Importance maps (top) are not enough for detecting such misalignments, environment variations are necessary.}
    \label{fig:misalignment}
    \vspace{-1mm}
\end{figure}

Deep reinforcement learning (RL) agents struggle to adapt to environments with slightly perturbed goal, while neurosymbolic (NS) agents learn isolated explicit skills, that can easily be combined or adjusted to adapt to environments' modifications.
Furthermore, learning agents are prone to shortcut learning, occurring when a model exploits superficial correlations in the training data, resulting in poor generalization to novel situations.
To identify if agents base their decision on the wrong input parts, eXplainable AI (XAI) methods, such as importance maps have been used~\citep{SchramowskiSTBH20,RasXGD22,RoyKR22,SaeedO23}.
In deep reinforcement learning (RL), such shortcut learning behavior is refereed to as goal misgeneralization~\citep{Koch2021-misalignment, Shah2022GoalMW, Tien2022CausalCA}, \ie to solving a sub-goal aligned with the \textit{true} objective. 
Such misalignments can be difficult to identify, as exemplified by~\cite{Langosco2022goal}, who showed that agents trained on Coinrun learn to run to the end of the level (sub-goal) instead of learning to reach the coin (true objective) (\cf Fig.~\ref{fig:misalignment}, Left) and by~\cite{delfosse2024interpretable}, with agents trained on Pong that learn to follow the enemy's paddle position instead of the ball's one (\cf Fig.~\ref{fig:misalignment}, Right). 
They show that hiding the enemy or changing its policy (\ie preventing it to move after it returns the ball) lead to agents incapable of catching the ball. Importance maps highlight the player's and enemy's paddles, and the ball, luring external observers into thinking that the agents ``understood'' that it needs to return the ball behind the enemy's paddle.
Map-based explanations indicate the importance of an input element for a decision without indicating \textit{why} this element is important, or how it is used within the decision process~\citep{Kambhampati2021SymbolsAA, rightconcepts, TesoASD23}.

The lack of generality of deep RL agents has previously been identified by~\cite{cobbe2020procgen}, who introduced Procgen, a set of procedurally generated environments, with varying assets (for the backgrounds and depicted objects) and level designs. 
The previously mentioned Coinrun game, with misaligned agents is part of Procgen, showcasing that varying the assets and procedural generation might not be enough.
The authors also underline that the Atari Learning Environments~\citep{bellemare2013arcade} is the by far most used RL benchmark. This test bench completely dominate the test of deep RL agents (\cf Fig.~\ref{fig:ale_vs_rest} in the Appendix), as it incorporates a diverse set of challenges, does not require extensive computations, nor suffer from any experimenter bias. 
However, Atari games do not provide any variations, making it impossible to test for generality or misalignment.

In this work, we propose \textbf{\textit{HackAtari}}, a framework that introduces novelty to the Atari Learning Environments. 
HackAtari contains a set of in total $50$ variations on $16$ 
Atari Learning Environments.
We show that these variations can be used to test identify RL agents' potential misgeneralizations, as well as to help in curriculum learning settings, by introducing simpler and more complex versions.

To learn robust policies, recent interpretable algorithms explicitly separate the extraction of neurosymbolic states from raw, pixel-based inputs~\citep{lin2020space,delfosse2021moc,zhao2023fast} from the action selection process. This allows for transparent action selection process, based on \eg first order logic~\citep{Delfosse2023InterpretableAE}, on concept bottlenecks~\citep{delfosse2024interpretable}, or on polynomial equations, later explained in natural language by an LLM~\citep{luo2024insight}. 
To efficiently train these methods, HackAtari incorporates the Object-Centric Atari framework (OCAtari)~\citep{delfosse2023ocatari}, that provides object-centric representations of the games, to allow for training and comparing both deep and interpretable NS agents. 
HackAtari also allows for the development of Continual RL method, as many environments provide different level of difficulty, based on the mastering of different skills, each trainable in a curriculum learning fashion. 
Specifically, our contributions are:
\begin{itemize}[leftmargin=20pt,itemsep=1pt,parsep=1pt,topsep=1pt,partopsep=1pt]
    \item[(i)] We introduce HackAtari, a set of modifications applied to different Atari environments, that allow for testing variations of the games. It allow to test generalization capabilities of RL agents, as well as training agents in curriculum learning settings.
    \item[(ii)] We demonstrate that our environments can be used to test already trained agents, as well as to train other agents on our variations.
    \item[(iii)] We use HackAtari to show the misgeneralisation of existing RL agents.
    \item[(iv)] We use show that HackAtari allows for learning LLM-defined reward functions.
\end{itemize}

We start off by introducing the HackAtari framework. We experimentally evaluate the generalization and learning capabilities of RL agents. Before concluding, we touch upon related work. 

\newpage

\section{HackAtari: Altered Atari Environments}
\label{sec:Hackatari}
In this section, we first explain how to create variations of the Atari Learning environments, then 
describe the $4$ alteration categories, that can be used to train and test different RL agents' capabilities.

\subsection{HackAtari: step, reset and reward modifications}
\begin{wrapfigure}{r}{0.45\linewidth}
\vspace{-4.5mm}
    \centering
    \includegraphics[width=.98\linewidth]{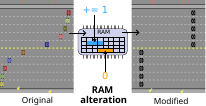}
    \caption{\textbf{RAM alteration allows for modified environments}. Exemplified on Freeway. Altering the some RAM cells leads to \textcolor{orange}{color} and \textcolor{cyan}{speed} changes.}  
    \vspace{-3mm}
    \label{fig:changeRAM}
\end{wrapfigure}

While being -by far-, the most used benchmark for training and testing RL agents, the Atari Learning environments do not openly provide any source code. 
We thus cannot directly modify the source code of these environments. 
However, \citet{Anand19AtariARI} have identified that the many useful information about the depicted objects can be observed in the RAM, and proposed \textit{AtariARI}, a wrapper that augments the info dictionary with pointers to the parts of the RAM responsible for \eg the position of the agent.
Inspired by this work, \citet{delfosse2023ocatari} created \textit{OCAtari}, a set of augmented Atari environments that directly provide neurosymbolic (object-centric) states of the games. 
They identify which part of the RAM are responsible for the object's properties (\eg visibility, position, color, orientation). 
They are thus able to provide both RGB and neurosymbolic states.

\begin{figure}[t]
    \centering
    \includegraphics[width=\linewidth]{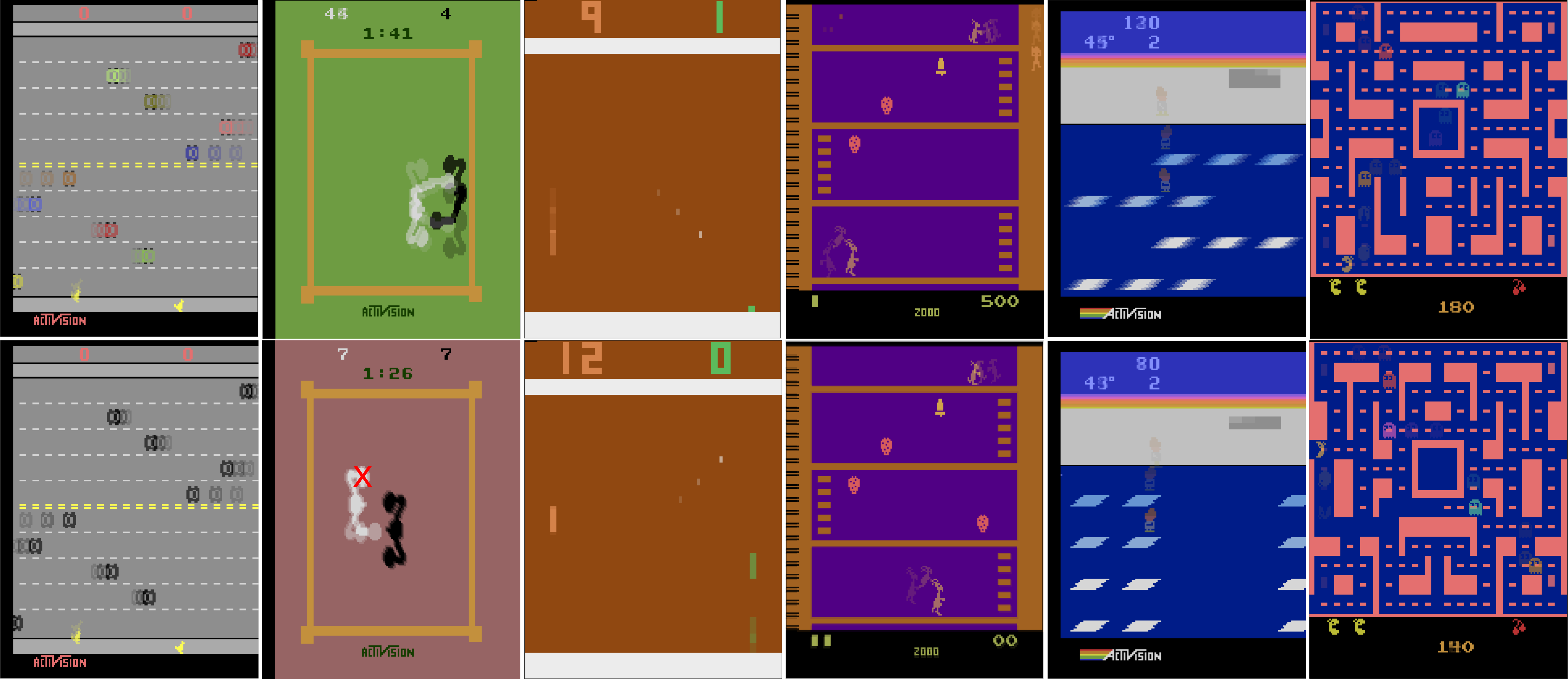}
    \caption{\textbf{HackAtari provides variations of Atari environments.} These include color changes (Freeway and Boxing), gameplay shifts (Boxing, MsPacman), continual learning settings (Kangaroo and Frostbite). The original games (top) are compared to HackAtari's modified versions (bottom). Superposed frames show the game dynamics.}
    \label{fig:multiple-examples}
\end{figure}

In our work, we use the mappings of RAM values to object states to understand the inner working of the game. 
We then can alter these RAM values to modify the behaviors of the objects. This is depicted in Fig.~\ref{fig:changeRAM} on \textit{Freeway}. 
For this game, the cars colors are encoded at the RAM cells \ramcell{77} to \ramcell{86}. By setting each of their values to $0$, we change the colors of each car to black. 
Furthermore, we identified the RAM position responsible for the movement of each cars (from \ramcell{33} to \ramcell{43}). 
By setting all of them to the same number, we can modify the speed of all the cars, such that they \eg now drive in columns. 
For some game, some object's properties can be encoded at multiple RAM cells. 
This is the case for the enemy's paddle position in \textit{Pong}, encoded both at RAM cells \ramcell{21} and \ramcell{50}. 
If we overwrite the RAM cell \ramcell{50} (to modify the behavior of the enemy's paddle, as shown in Fig.~\ref{fig:multiple-examples}), the value will not be used by the program and will be overwritten at the next time step by the game. 
Altering the RAM cell \ramcell{21} allows for modifying the enemy's policy.

Modifying \textit{Pong} to obtain a \textit{Lazy Enemy Pong} version is more complicated than the previously illustrated changes. 
Our goal is to have the enemy static after it touches the ball to return it (and a moving enemy after the agent returned the ball). 
For this, we need to keep track of the position of the ball and of the enemy. By comparing the ball's positions at the previous and current timestep, we can decide whether to enforce the enemy to remain static, or not. 
If the ball is going to the agent's side, we overwrite the enemy's position to the one it had when it touched the ball.

Our framework incorporates several modification types, that can be combined (\eg altering the cars' color and the cars' speed on \textit{Freeway}). To make these modifications, $3$ types of functions can be used:
\begin{enumerate}[leftmargin=20pt,itemsep=1pt,parsep=1pt,topsep=1pt,partopsep=0pt]
    \item[(i)] \textit{state modification at each step}, \eg when the position of an object is being altered, or when objects are being disabled (\ie the corresponding RAM cells are constantly being overwritten).
    \item[(ii)] \textit{state modification at reset time}, \eg when the level layout is being swapped, or when the agent is spawning at a randomized initial position.
    \item[(iii)] \textit{reward modification} (at each step), when reward is changed to modify the goal in the environment, such as encouraging peaceful policies in shooting games (or discouraging the use of AMO).
\end{enumerate}

We have explained how to alter the RAM to obtain a resource efficient way of altering the learning environments. For implementation details on other environments, please look at our open-source repository, that we release together with this paper. In the following, we explain how we categorize the environments' modifications based on the type of training and/or testing that they allow for.

\newpage
\subsection{Testing visual and dynamics robustness, curriculum RL and adaptability.}
\label{sec:paradigm}

\begin{wraptable}{r}{0.45\textwidth}
  \vspace{-4.4mm}

\newcommand{\rtexttt}[1]{\rotatebox[origin=c]{90}{\texttt{#1}}}

\newcommand{\rtexbttt}[1]{\rotatebox[origin=c]{90}{\texttt{#1}}}
 
\setlength{\tabcolsep}{1pt}
\centering
\caption{\textbf{HackAtari variations.} $\checkmark$ mark at least one existing variation of the environment within HackAtari.}

\resizebox{.45\textwidth}{!}{
\begin{tabular}{c|ccccccccccccccccccc}
\multicolumn{1}{c|}{\rotatebox[origin=c]{90}{\makecell{Modification}}} &
\multicolumn{1}{l}{\rtexbttt{Bankheist}} &
\multicolumn{1}{l}{\rtexbttt{BattleZ.}} &
\multicolumn{1}{l}{\rtexbttt{Boxing}} &
\multicolumn{1}{l}{\rtexbttt{Breakout}} &
\multicolumn{1}{l}{\rtexbttt{Carnival}} &
\multicolumn{1}{l}{\rtexbttt{ChopperC.}} &
\multicolumn{1}{l}{\rtexbttt{DonkeyK.}} &
\multicolumn{1}{l}{\rtexbttt{FishingD.}} &
\multicolumn{1}{l}{\rtexbttt{Freeway}} &
\multicolumn{1}{l}{\rtexbttt{Frostbite}} &
\multicolumn{1}{l}{\rtexbttt{Kangaroo}} &
\multicolumn{1}{l}{\rtexbttt{Montezuma.}} &
\multicolumn{1}{l}{\rtexbttt{MsPacman}} &
\multicolumn{1}{l}{\rtexbttt{Pong}} &
\multicolumn{1}{l}{\rtexbttt{Riverraid}} &
\multicolumn{1}{l}{\rtexbttt{Seaquest}} &
\multicolumn{1}{l}{\rtexbttt{Skiing}} &
\multicolumn{1}{l}{\rtexbttt{SpaceInv.}} &
\multicolumn{1}{l}{\rtexbttt{Tennis}} \\
\midrule
Color &
& 
&
$\checkmark$& 
&
&
&
&
&
$\checkmark$& 
$\checkmark$& 
&
&
&
&
&
&
&

\\
Gameplay &
$\checkmark$&
$\checkmark$&
$\checkmark$&
$\checkmark$&
$\checkmark$&
$\checkmark$&
$\checkmark$&
$\checkmark$&
$\checkmark$&
$\checkmark$&
$\checkmark$&
$\checkmark$&
$\checkmark$& 
$\checkmark$&
$\checkmark$&
$\checkmark$&
$\checkmark$&
$\checkmark$&
$\checkmark$
\\
Skill &
$\checkmark$&
$\checkmark$&
&
&
$\checkmark$& 
$\checkmark$&
$\checkmark$&
$\checkmark$&
$\checkmark$&
$\checkmark$&
$\checkmark$&
&
$\checkmark$&
$\checkmark$&
$\checkmark$&
&
&
&
$\checkmark$
\\
Goal &
&
&
&
&
&
&
$\checkmark$&
$\checkmark$&
&
&
&
&
$\checkmark$&
&
$\checkmark$&
$\checkmark$&
&
$\checkmark$&
\\
\bottomrule

\end{tabular}}
\label{tab:supported games}
 
  \vspace{-4mm}
\end{wraptable}

HackAtari environments alterations aim at testing different capabilities of RL agents. 
We thus categorize these modifications into the $4$ following paradigms.

\textbf{1. Visual Domain Adaptation.} Altering the RAM allows changing the colors of the depicted objects in many environments. This is showcased in Fig.~\ref{fig:changeRAM} on \textit{Freeway}. 
These shifts can be used to test simple shortcut learning. 
For this type of shortcut learning, neural networks learn to associate specific pixel values to actions, as identified by ~\cite{stammer2021right} on classification tasks. 
We also provide a general way of modifying any pixel color for any ALE game, exemplified in Fig.~\ref{fig:multiple-examples} on \textit{Boxing}, using simple pixel values mapping and replacement.
HackAtari incorporates a testing script, that automatically randomly swap the present pixel values.
We however advise RL practitioners to use these settings to test their agents only, or to build offline RL datasets, as the color processing of the frame is resource intensive. 
For faster execution, we recommend using the color swaps done by our RAM alterations techniques for training agents. 
We mark them in the \textit{Color} line of Tab.~\ref{tab:supported games}.

\textbf{2. Dynamics Adaptation.} For this changes, we alter the gameplay, and agents have to adapt to small perturbations. 
These games can be used to test the robustness and adaptability of RL agents. 
Examples of gameplay shifts are: small drifts in ball games (\eg \textit{Breakout} and \textit{Tennis}), gravity slightly pulling the agent downwards (in \eg Boxing, Pong, Seaquest), level layout change \eg in \textit{MsPacman} (\cf Fig.~\ref{fig:multiple-examples}), velocity modifications (\eg on cars in \textit{Freeway}, on the projectiles in \textit{Carnival}). 
For this category of games, there is no distribution shift in term of pixel values, but the agents have to adapt to novel situations. 
They also allow to detect another shortcut learning type, such as the one described in introduction on Pong, where the agent learns to follow the enemy instead of the ball~\citep{delfosse2024interpretable, kohler2024interpretable}. 
HackAtari incorporates a version of \textit{Pong}, where the enemy is static after returning the ball (\cf Fig.~\ref{fig:multiple-examples}).
Some games, such as \textit{MsPacman}, \textit{Kangaroo}, \textit{Montezuma's Revenge} also include multiple levels or level parts, that are usually accessible if the first level is completed. For these, we provide one type of modification that randomly select one level/part after each death, or each reset of the environment. This allows the agent to learn more generalizable policies (or to be tested), for abilities such as navigate another maze layout, \cf \textit{MsPacman} in Fig.~\ref{fig:multiple-examples}).

\textbf{3. Curriculum Reinforcement Learning (CRL).} In this learning paradigm, the task complexity is incrementally increased. CRL environments are used to assess the agents' ability to learn different skills in a structured manner with incremental complexity. For example, we created versions of \textit{Kangaroo} with disabled monkeys and disabled thrown coconuts, of \textit{Seaquest} without enemy and with unlimited oxygen, or Freeway with stopped cars. 
These can be used to separately learn navigations from other skills~\citep{Mirowski2016LearningTN, Sharifi2023TowardsSA}, or to learn different options (\ie high level action) using small networks~\citep{stolle2002learning, BaconHP17}. Novel research is also looking into neural merging~\citep{Kirkpatrick2016OvercomingCF, Brahma2021HypernetworksFC}, lately brought to continual RL~\citep{gracla2023multi}. Finally, neurosymbolic methods often explicitely learn separated skills~\citep{Kimura21NeSyRL, Delfosse2023InterpretableAE} on their logic-based RL agents.

\textbf{4. Reward Signal Adaptation.}  
We also incorporate an easy reward modification implementation, to test the agent's ability to rapidly adapt to new objectives, potentially using large language models (LLMs) to generate the reward signals~\citep{ma2023eureka, Xie2023Text2RewardRS}, or to evaluate how well RL agents align with human societal values~\citep{pan2021effects}. 
HackAtari enables users to specify rewards defined by large language models (LLMs) and to incorporate alternative reward structures that change the game's objectives. For example, in the game Kangaroo, the agent can be rewarded more for punching monkeys than for saving the joey. By reducing the reward for aggressive actions and increasing it for reaching the joey, a different policy is created that is more aligned with the original game's goals~\citep{delfosse2024interpretable}. More broadly, this approach can be used to decrease incentives for aggressive behaviors, such as in shooting games~\citep{pan2021effects}, or to leverage LLMs for generating various reward types in skill acquisition tasks.
A comprehensive example of modifying and adjusting the reward function is provided in App.~\ref{app:changingtheRF}.

\section{Experimental Evaluation}
\label{sec:experiments}

In our evaluation, we investigate several benefits of using HackAtari over the original Atari Learning Environments. Specifically, we aim at answering the following research questions:
\begin{itemize}[itemsep=2pt, parsep=0pt, topsep=0pt, partopsep=0pt]
    \item[\textbf{(Q1)}] Are our HackAtari variations usable for training RL agents?
    \item[\textbf{(Q2)}] Can HackAtari variation reveal misalignment and be used to correct it?
    \item[\textbf{(Q3)}] Can alternative reward functions allow for adjusting learned behaviors?
    \item[\textbf{(Q4)}] Can some HackAtari environments be used for curriculum learning?
\end{itemize}
\textbf{Experimental Setup.} 
We compare PPO~\citep{Schulman2017ProximalPO} and C51~\citep{BellemareDM17} agents' on our modified environments to ones trained on the original ones (all averaged over $3$ seeds). We also let human agents train on the original environments, and test them on both the original and modified versions (\cf App.~\ref{app:human_eval} for details on this user study). We average human scores across at least $3$ human users. 
All agents use the classical DQN input representation~\citep{Mnih2015dqn}, \ie process $4\!\times\!84\!\times\!84$ frame stack and are trained on $10$M frames, on a $40$ GB DGX A100 server. 
We used PPO implementation of~\citet{huang2022cleanrl} with its default hyperparameters (\cf App.~\ref{app:hyperparameter}). 
We focus the evaluations on robustness of the used policy, rather than achieving the best possible scores. 

\textbf{Game variations details.}
We here concisely present the games' variations that we refer to in this section. 
A more detailed list can be found in App.~\ref{app:variants1} and a list of all modifications in App.~\ref{app:modifications}.
\\\textbf{One Armed (\textit{Boxing}):} The agent can only punch its opponent with its right arm (instead of both).
\\\textbf{No Barrel (\textit{DonkeyKong}):} We cancel falling barrels that the agent must avoid.
\\\textbf{Aligned Cars (\textit{Freeway}):} All cars have identical speed, and thus travel aligned. 
\\\textbf{Mono-Colored (\textit{Freeway}):} All the cars have identical colors (\eg black, \cf Fig.~\ref{fig:multiple-examples}).
\\\textbf{Stopped Cars (\textit{Freeway}):} All the cars are visible but static, allowing any agent to safely cross.
\\\textbf{No Danger (\textit{FrostBite}):}  The horizontally traveling enemies (in between the ice blocks) are canceled. 
\\\textbf{Static Ice (\textit{FrostBite}):}  The ice blocks are static instead of moving sideways (\cf Fig.~\ref{fig:multiple-examples}). 
\\\textbf{No Danger (\textit{Kangaroo}):} We cancel both monkey enemies and deadly falling coconuts.
\\\textbf{Safe and Close (\textit{Kangaroo}):} No Danger + The agent randomly spawns on $1^{st}$, $2^{nd}$ or $3^{rd}$ floor.
\\\textbf{Swap Level (\textit{MsPacman}):} The agent spans in a level with a different maze layout (\cf Fig.~\ref{fig:multiple-examples}).
\\\textbf{Lazy Enemy (\textit{Pong}):} The enemy stands still after returning the ball instead of following it (\cf Fig.~\ref{fig:multiple-examples}). 
\\\textbf{Infinite Oxygen (\textit{Seaquest}):} The level of oxygen does not decrease (stays at $100$\%).
\\\textbf{No Shield (\textit{SpaceInvaders}):} Shield that cancel missiles from the agents and the enemies are removed.

\begin{figure}[t]
    \centering
    \includegraphics[width=1.\linewidth]{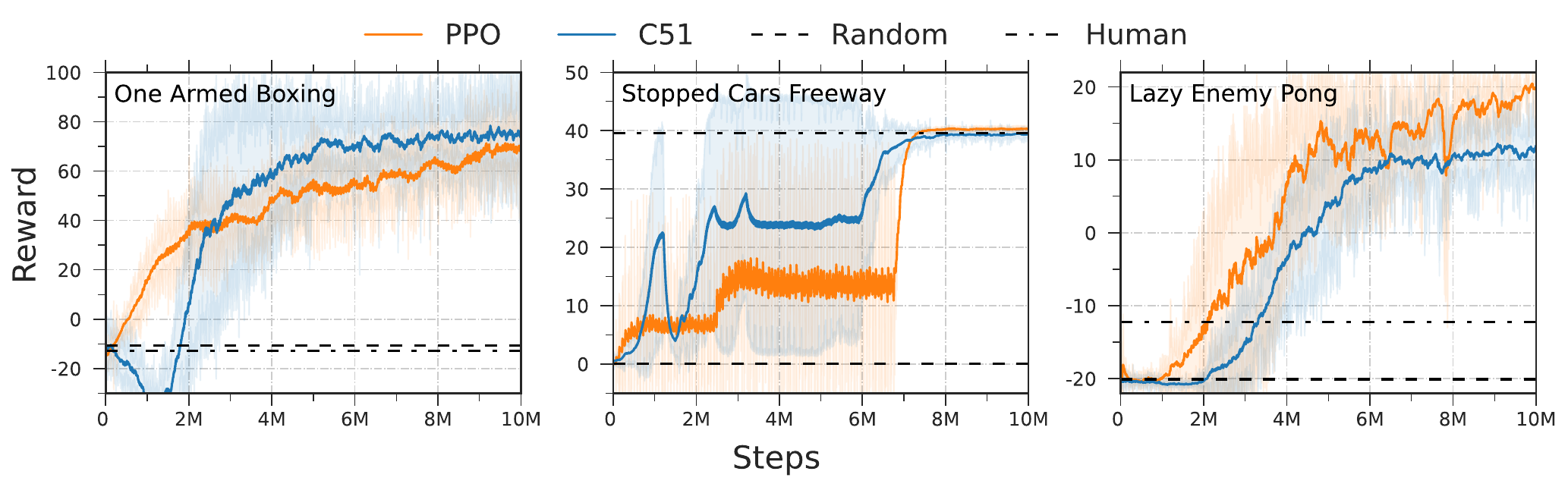}
    \vspace{-2mm}
    \caption{\textbf{RL agents can learn on altered environments}, exemplified on \textit{One armed Boxing}, \textit{Mono-Colored Freeway} and \textit{Lazy Enemy Pong}, by PPO and C51 agents. These agents are able to progressively improve from random to (or beyond) the human level.
    \textit{Freeway}'s high variance is due to the number of frames needed before each seeded agent reaches the top.}
    \label{fig:learnable}
\end{figure}

\paragraph{HackAtari modified environments can be used for learning (Q1).} Before other investigations, we need to verify that our environments are usable to train RL agents, \ie that they do not lead to games impossible to complete.
After training on our modified versions, PPO and C51 agents consistently demonstrate the ability to master the introduced changes (\cf learning curves in Fig.~\ref{fig:learnable}, PPO vs. Random in Tab.~\ref{tab:robustness1} and the additional curves in App.~\ref{app:additional_results}), \ie are able to increase their scores in our variants. 
Our findings confirm that these game variants can be used to train policies and evaluate the generalization capabilities of agents trained on the original environments (to \eg compare them to humans or agents directly trained on the variations).

\paragraph{HackAtari can help uncover flaws of  trained agents (Q2).} 
Let us now test the generalization capabilities of artificial agents using HackAtari's variations. 
PPO's performances remarkably drop on all of our tested variations, whereas humans' ones increase in all but \textit{Boxing} (\cf Tab.~\ref{tab:robustness1}). For this game, the PPO agents trained on this variation are able to adapt their policies and achieve comparable performances, even with the absence of the second arm.
For \textit{Kangaroo} and \textit{DonkeyKong}, the lack of enemies/barrels simplifies the game, allowing humans to safely reach the end of the level, a behavior that PPO agents have not learned, as they were not able to observe reward for this behavior during this training. 
These results show that autonomous agents could benefit from incorporating other learning or planning modules, \eg (vision) language models, to guide the agent in constantly changing environments.
We can also confirm the findings of~\cite{delfosse2024interpretable}: RL agents trained on \textit{Pong} learned misaligned behavior, heavily relying on the enemy's position (instead of the ball's one). 
Rearranging the walls also lead to performance drops in \textit{MsPacman}, showing that these agents follow a remembered pattern instead of learning navigation~\citep{sukhbaatar2018intrinsic, burda2018exploration}.

\begin{table}[b]
\centering
\caption{\textbf{Deep RL agents' performances drop when confronted with slightly novel situations.} PPO agents scores, trained and/or evaluated the original and variation of different Atari games. 
Gameplay or color changes mostly lead to performance drop, whereas most human scores increase.
} 
\resizebox{1.0\textwidth}{!}{
\begin{tabular}{@{}l|rrr|rr|r@{}}
\toprule
\textbf{Game}          & \multicolumn{3}{c|}{\textbf{PPO}}             & \multicolumn{2}{c|}{\textbf{Human}} & \textbf{Random} \\ \midrule
\textbf{Training}      & original        & original        & variation & original          & original         & -               \\
\textbf{Testing}       & original        & variation       & variation & original         & variation        & variation       \\ \midrule
\textbf{Boxing (OA)}        & $90.9\smallpm{1.5}$ & $1.9\smallpm{10.2}$  &  $82.2\smallpm{9.3}$         & $0.6\smallpm{2.7}$      & -$12.8\smallpm{18.8}$  & -$10.8\smallpm{0.9}$                 \\
\textbf{DonkeyKong (NB)}       & $3480\smallpm{1032}$  & $0\smallpm0$  & $0\smallpm0$          & $7320\smallpm{3961}$  & $50000\smallpm{0}$ & $0\smallpm0$                \\
\textbf{Freeway (AC)}       & $31.4\smallpm{1.5}$  &  $20.4\smallpm{0.7}$  & $29.1\smallpm{1.8}$          & $21.7\smallpm{4.8}$    & $22.4\smallpm{1.6}$    & $0\smallpm{0}$                \\
\textbf{Freeway (MC)}       & $31.4\smallpm{1.5}$  &  $24.6\smallpm{2.7}$  & $32.7\smallpm{0.8}$ &  $21.7\smallpm{4.8}$  & $29.3\smallpm{1.5}$ & $0\smallpm{0}$             \\
\textbf{Frostbite (SI)}       & $313\smallpm{13.1}$ & $265\smallpm{25.4}$  & $991\smallpm{390}$   & $4916\smallpm{3278}$  & $29360\smallpm{19120}$   &  $59.4\smallpm{43}$               \\
\textbf{Kangaroo (ND)}      & $1838\smallpm{650}$ &  $0\smallpm{0}$               &  $0\smallpm0$         & $2344\smallpm{1434}$ & $12200\smallpm{1555}$   & $0\smallpm0$                \\
\textbf{MsPacman (SL)}      & $2312\smallpm{465}$ & $456\smallpm{260}$ & $2228\smallpm{428}$          & $4592\smallpm{3725}$ & $6149\smallpm{5097}$ &  $135\smallpm{65}$               \\
\textbf{Pong (LE)}          & $16.0\smallpm{3.4}$  & -$12.6\smallpm{2.4}$ &  $18.1\smallpm{4.4}$         & -$13.7\smallpm{2.3}$   & -$12.2\smallpm{6.4}$       & -$20.1\smallpm{0.4}$                 \\
\textbf{SpaceInv. (NS)}     & $724\smallpm{123}$                &  $496\smallpm{78}$               &  $1181\smallpm{292}$   & $640\smallpm{368}$   & $726\smallpm{616}$   & $109\smallpm{32}$                \\ \bottomrule
\end{tabular}
}
\label{tab:robustness1}
\end{table}

\newpage
\begin{wrapfigure}{r}{0.39\textwidth}
  \vspace{-1mm}
  \includegraphics[width=\linewidth]{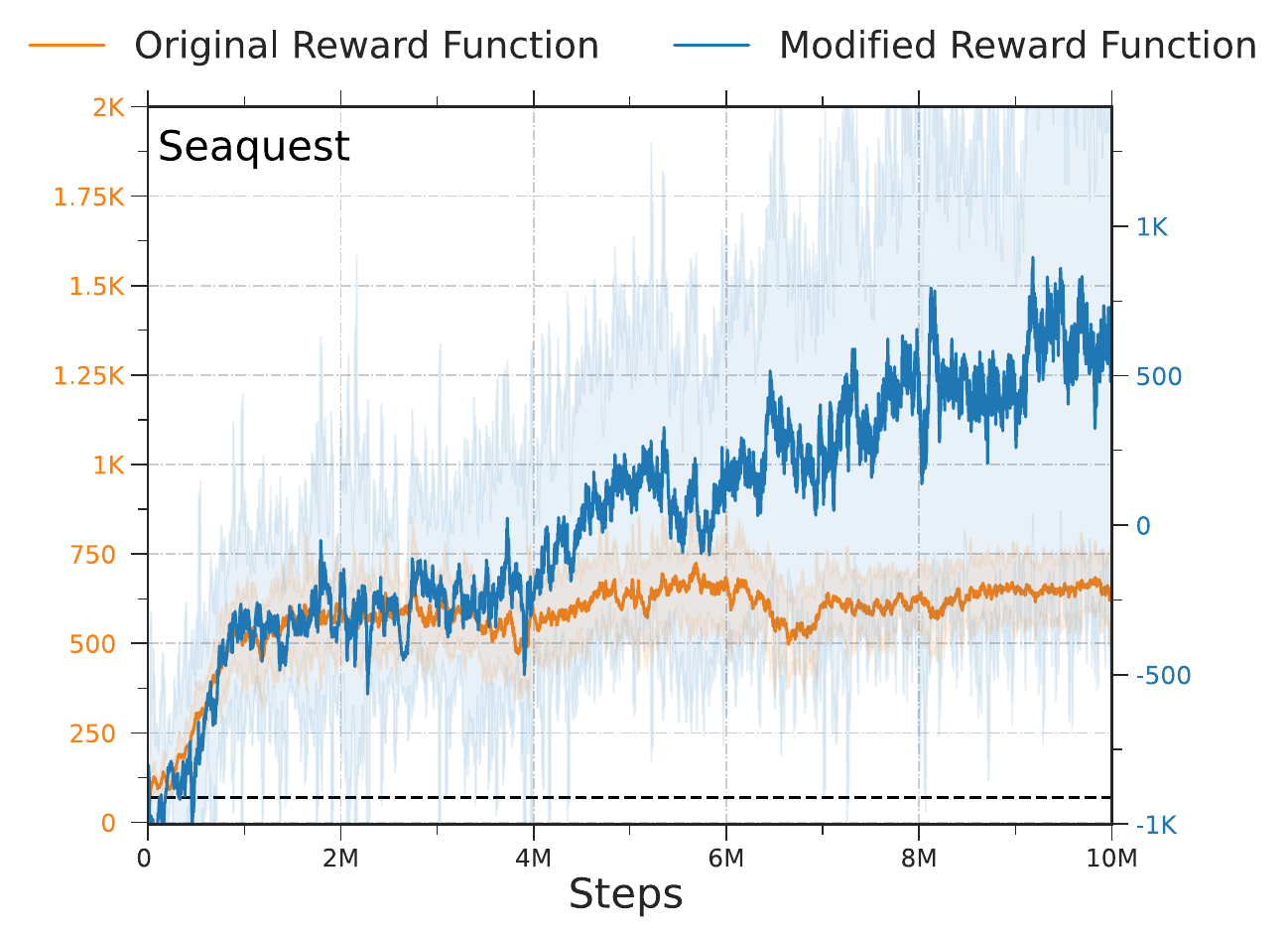}
  \vspace{-5mm}
  \caption{\textbf{LLM can guide RL agents.} Performances of PPO agents trained using an LLM-provided reward function (blue) and the original reward (orange).}
  \label{fig:rfchanged}
  \vspace{-3mm}
\end{wrapfigure}

\paragraph{HackAtari allows to learn alternative behaviors (Q3).}
We make use of the integration of alternative reward function of HackAtari to test if agents can learn based on \eg LLM provided guidance (detailed in App.~\ref{app:changingtheRF}) instead of the original score, included in the ALE games. 
The original Seaquest game does not reward agents for saving divers, but only for shooting enemies. 
In this experiment, we encourage this more pacifist behavior by rewarding for rescuing them. 
The LLM provided us with a reward signal aligned with this goal (\cf App.~\ref{app:changingtheRF}), that we used to train PPO agents (blue line in Fig.~\ref{fig:rfchanged}). The LLM was able to produce a useful reward function, allowing the agents to learn.
Similarly, expert defined functions can be used, as done by~\citet{delfosse2024interpretable}, who redefine reward signals in Pong (adding a penalty on distance between the ball and the player), Kangaroo (rewarding the progress of the mother kangaroo towards its joey), and on Skiing (fixing the reward ill-defined reward, by directly rewarding the agents when they pass in between the poles). We were able to reproduce their results by integrating their provided reward function to the HackAtari environments.

\begin{wrapfigure}{r}{0.39\textwidth}
  \vspace{-5mm}
  \includegraphics[width=\linewidth]{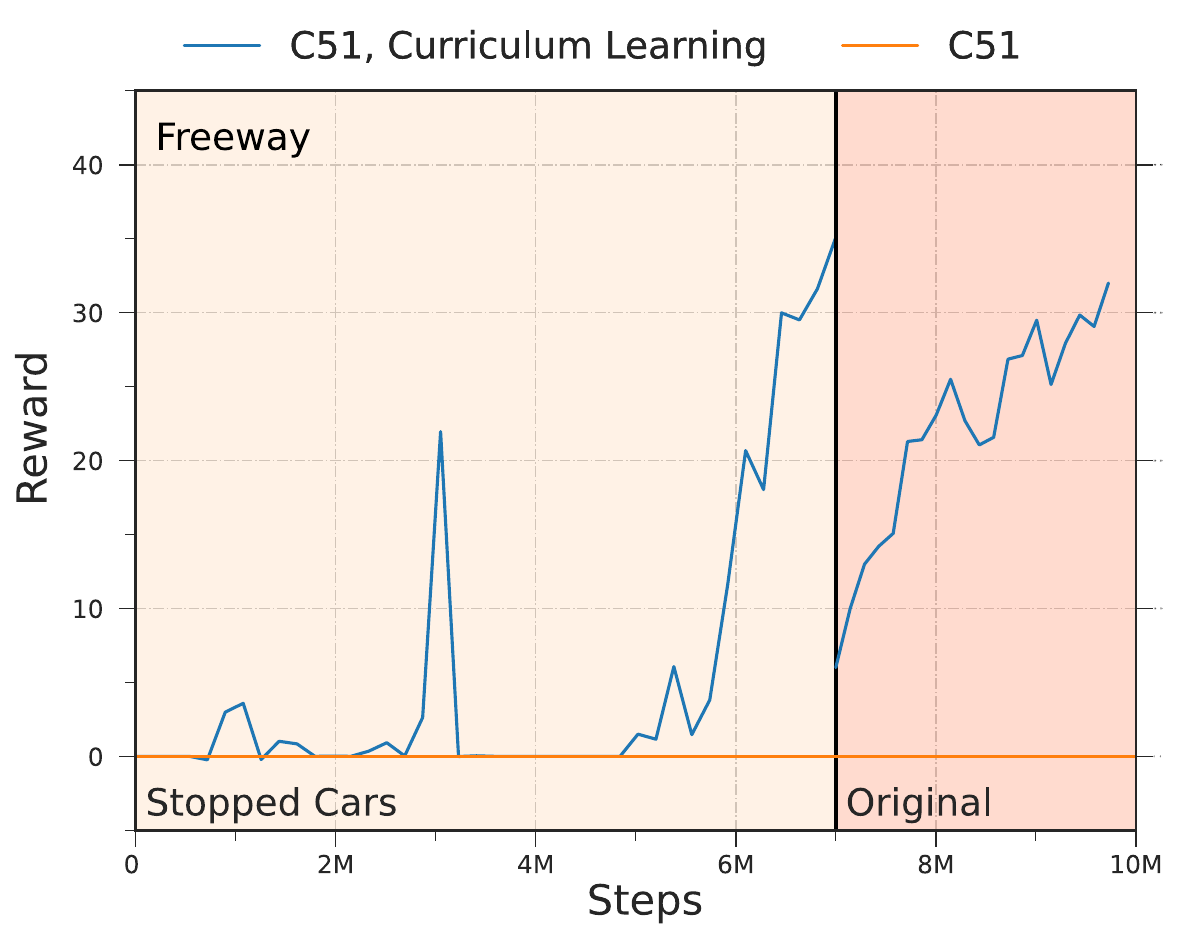}
  \vspace{-5mm}
  \caption{\textbf{Curriculum learning, using a simplified version.} C51 agent learn to reach the top (for $7M$ steps on easier variant), before learning to avoid cars.}
  \label{fig:curriculum}
  \vspace{-3mm}
\end{wrapfigure}
 
\paragraph{Simplifications enable skill learning (Q4).}

\begin{figure}[b]
    \centering
    \vspace{-2mm}
    \includegraphics[width=\linewidth]{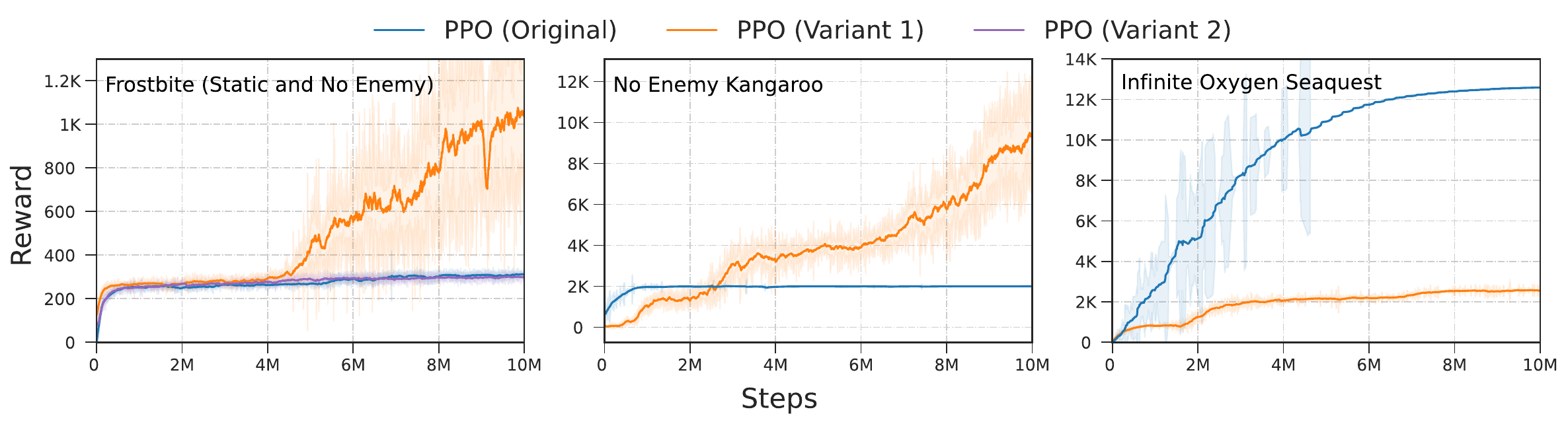}
    \caption{\textbf{Alternate versions allow to learn different skills:} (i) In \textit{Frostbite}, RL agents can learn to safely collect ice blocks and fishes on the \textit{No Danger} variant, then to avoid the enemies on a simpler \textit{Static Ice} second variant (left), (ii) with \textit{Safe and Close} agents learn to reach the joey, while they learn to punch Monkey in the original game (center), (iii) in \textit{Infinite Oxygen Seaquest}, agents can also learn to kill enemies (as in the original), but also to collect divers' group (right).}
    \label{fig:skill_learning}
    \vspace{-2mm}
\end{figure}

As shown in Tab.~\ref{tab:robustness1}, agents are able to learn the game of \textit{Freeway} within $10M$ steps. However, in our experiment, we observed that C51 agents fail to do it if we do not use the frame skipping optimization (slightly increasing the sparsity). 
To observe reward, the randomly playing agent needs to cross the $10$ lines, knowing that it is pushed back down when hit by a car.
Using our \textit{Stopped Cars} variation of \textit{Freeway}, we decrease this sparsity, allowing the agents to observe reward in less than a million steps. 
We were thus able to first train the agent for $7M$ steps on this simpler variation, allowing it to learn that reaching the top provides reward, before continuing learning for the last $3M$ steps, that allow it to adjust to avoid cars (\cf Fig.~\ref{fig:curriculum}). 
These experiments showcase that HackAtari gameplay simplifications can be used to now evaluate curriculum RL techniques on ALE.\\
Our next experiments demonstrate the usability of HackAtari for parallel skills' acquisition. \\
In \textit{Frostbite}, the agent has to collect ice blocks to build its igloo (\cf Fig.~\ref{fig:multiple-examples}), by jumping on moving ice platforms, while avoiding enemies. It can also collect fishes, which provides additional reward. 
Collecting blocks without falling in the water can be more easily learned, using our first \textit{No Danger} variant, while
the enemy-aware navigation skill can be acquired on the second \textit{Static Ice} variant (\cf Fig.~\ref{fig:skill_learning}, left). 
Merging these overlapping skills could be achieved using neural merging techniques.\\
In \textit{Kangaroo}, ~\cite{delfosse2024interpretable} identified that agents learn to punch monkeys instead of saving the joey, as this originally intended goal is more difficult to reach, and use an alternate reward signal to further incentivize for saving the joey. 
We here identify that fighting monkeys leads to $4.5$ times higher scores than the goal described in the game manual\footnote{\mbox{\url{www.retrogames.cz/play_195-Atari2600.php}}} (\cf Fig.~\ref{fig:skill_learning}, center). Indeed, the agents learn to reach the joey on our \textit{Safe and Close} variation, that cancels the monkey enemies. 
Interpretable skill-based agents could learn these two behaviors in a curriculum learning manner, as well as when to select each, and alternate reward signals could also boost the learning speed of the saving behavior.\\
Finally, in \textit{Seaquest}, the agents usually learn to shoot enemies until their oxygen bar is depleted. Our \textit{Infinite Oxygen} version allows it to learn to kill more enemies, as well as surfacing when enough divers have been saved  (\cf Fig.~\ref{fig:skill_learning}, right).
We also provide a way to cancel the enemies (even if not used here). Again, this demonstrates that HackAtari's gameplay simplications can allow for controlled separate skill learning (or options~\citep{BaconHP17}). 
Other environments, such as \textit{MsPacman} with \textit{Caged Ghosts}, could allow for first learning to navigate the different mazes, before releasing the ghosts and have the agent learning to adapt its behavior to avoid getting killed by them (or to collect the superpower pills that allow to chase them). We leave this for future work.

Overall, our experimental evaluations have demonstrated that HackAtari can be used to train and evaluate agents on alternate tasks within the same (or similar) environment, for which humans require no retraining. 
We also demonstrated that they can be used to emphasize the acquisition of alternative skills, or of curriculum learning (on simplified environments \eg \textit{Stopped Cars Freeway}).

\section{Related Work}
\paragraph{Novelty in RL.} In general, novelty can be attributed to the characteristics of an object, to an event involving the object, or to an act of manipulation with an object. Although there is no strict definition of novelty in RL or ML, it often relies on either novelty detection~\citep{MarkouS03, MarkouS03a, Schmidhuber08, PIMENTEL2014215} or intrinsic-motivation-based exploration~\cite{OudeyerK09, SinghLBS10, barto2013novelty, SiddiqueDRM17}. The latter, as described by \cite{harlow1950learning}, describes intrinsic motivation as the drive to manipulate and explore features, \ie explore uncertain or novel elements of the environment. While HackAtari works as a benchmark or test suite for novelty detection, in this work, we focus on the aspect of HackAtari as a valuable tool for identifying and addressing misalignment within reinforcement learning environments.

\paragraph{Evaluating Generalization and Robustness.} 
Assessing generalization and robustness gets increasing attention~\cite{tec2023space, linial2023benchmarks, busch2024truth}, particularly in RL, as it ensures that AI agents can apply their learned behaviors effectively. Recently, some started to develop special benchmarks to grapple with this in multiple ways \citep{Nichol18Sonic, justesen2018illuminating, JulianiKBHTHCTL19, CobbeKHKS19}. 
Procgen~\citep{cobbe2020procgen}, provides a standardized platform to evaluate generalization and robustness across diverse, procedural generated environments. This innovation addresses critical challenges, \eg, overfitting or studying transfer learning capabilities effectively and plays a pivotal role in advancing RL algorithms \citep{cobbe2020procgen, mohanty2021measuring}. Atari~\citep{bellemare2013arcade} on the other hand, as stated by \cite{cobbe2020procgen}, is the gold standard for benchmarking in RL but lacks variety in the state spaces to be used to evaluate generalization and robustness. 
While the first steps have been done in this direction \citep{Farebrother18generalization,tomilin2024coom}, with HackAtari, we start adding more of the needed variety to the ALE by generating new versions of already known games. 
Other work have looked into plasticity to help agents to adapt to change. Thus,~\cite{delfosse2021rationalrl} categorize Atari games based on their amount of change during the learning phase, but do not evaluate agents on unseen data.

\paragraph{Continual reinforcement learning benchmarks.}
Atari games serve as crucial testing grounds for increasingly complex RL methods~\citep{hessel2018rainbow,Hafner2020dreamer,Badia2020agent57,Fan2021GDI,farebrother2022proto,xuan2024rezero}. 
Despite achieving superhuman performance, challenges such as efficient exploration, algorithmic efficiency, planning with sparse rewards, sample inefficiency, and generalization failures persist. 
The need for expanded Atari benchmarks is recognized by researchers such as \citet{Toromanoff2019Saber} and \citet{Fan2021AtariBenchmark}, who propose additional metrics for accurate performance assessment. Further, efforts like the Atari $100k$ benchmark by \citet{KaiserBMOCCEFKL20}, curated subsets of Atari Learning Environment (ALE) environments by \citet{AitchisonSH23}, and the Mask Atari benchmark for POMDPs by \citet{Shao2022POMDP} contribute to advancing Atari as a benchmarking tool in RL research even further. 
HackAtari follow on from here and offers a wide range of applications, such as benchmarking for generalization, to help with explainable RL tasks or create new tasks for continuous RL approaches. 
We believe that comparison methods between different continual methods~\citep{mundt2021cleva} should be brought to RL.




\section{Limitations}
\label{sec:limitations}
While HackAtari provides a robust framework for evaluating RL agents, it is currently limited to the set of Atari Learning Environments. 
This restricts the ability to fully assess transfer learning and adaptability across a broader spectrum of tasks. 
The findings may not always generalize to more complex or varied environments, highlighting the need to extend HackAtari to incorporate a wider variety of scenarios, including 3D environments and more complicated real-time strategy games.

Additionally, we believe that further research is needed to integrate human learning principles more effectively into RL frameworks. 
Moreover, while large language models and vision models show promise for enhancing RL agents, they introduce complexities and require significant computational resources, posing challenges for seamless integration.
Addressing these limitations will be crucial for realizing the full potential of agents able to adapt to HackAtari variations in driving autonomous agents' innovation.
To better evaluate the gap between such agents and human ones, we are conducting a broader user study investigation, that involve more subjects on more game variations. 

\section{Conclusion}

In this study, we introduce HackAtari, a framework designed to test the generalization, robustness, and curriculum learning capabilities of RL agents on the most commonly used set of Atari Learning Environments. 
By offering a wide range of modifications to existing Atari games, HackAtari will enable researchers to create more human-like and adaptive agents, addressing key challenges.
It also serves as a valuable tool for studying the behavior and decision-making processes of RL agents, offering insights into how agents adapt to novel environments and tasks, and helping to uncover various shortcut learning behaviors, such as RL agents following the enemy on \textit{Pong}, or learning a navigation path on \textit{MsPacman}.
Through a series of experiments, we evaluate the efficacy of HackAtari in uncovering such misalignments, testing RL agents' adaptability to novel environments, and enabling curriculum learning in the Atari games. 


\section{Ethical Consideration and Broader Impact}
\label{sec:ethics}
The development and use of HackAtari raise ethical considerations. 
Researchers must be vigilant about the potential misuse of adaptive agents. 
Furthermore, while our variations are lightweight, the computational resources required for advanced models may have environmental impacts, necessitating efforts to develop more energy-efficient algorithms.
HackAtari also has potential for broader impacts. 
By facilitating the creation of more human-like and adaptable autonomous agents, it can drive advancements in various fields, from autonomous systems to interactive entertainment.
However, it is essential to balance innovation with ethical responsibility, ensuring that these technologies benefit society as a whole and do not exacerbate existing inequalities or create new ethical dilemmas.

\section*{Acknowledgment}
We would like to express our sincere gratitude to our students Felicia Benkert, Lars Wassmann, Mohammad Azimpour, Svenja Droll, Annabell Schmidt, Julian Mehler, Markus Lang and Tim Blaschzyk for dedication and feedback. Their enthusiasm and perseverance have been instrumental in achieving the results presented in this paper. We also thank our user study participants for their time and insights. Thank you for your continuous efforts and commitment.

This research work has been funded by the German Federal Ministry of Education and Research, the Hessian Ministry of Higher
Education, Research, Science and the Arts (HMWK) within their joint support of the National Research Center
for Applied Cybersecurity ATHENE, via the ``SenPai: XReLeaS'' project as well as their cluster project within the Hessian Center for AI (hessian.AI) ``The Third Wave of Artificial Intelligence - 3AI''. 

\bibliography{bib}
\bibliographystyle{plainnat}

\appendix

\section{On the impact of Atari games in RL.}

\begin{figure}[h!]
    \centering
    \includegraphics[width=0.55\linewidth]{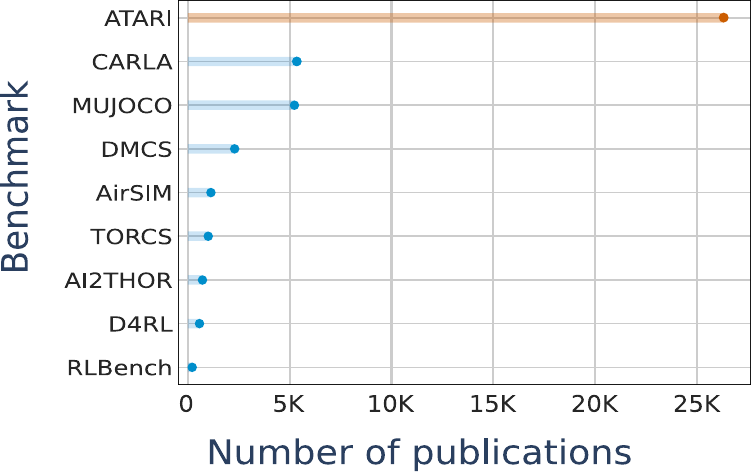}
    \caption{\textbf{RL research needs diverse Atari environments.} The Atari Learning Environments is, by far, the most used RL benchmark among the ones listed on \url{paperswithcode.com} .}
    \label{fig:ale_vs_rest}
\end{figure}

\section{Details on and Additional Results}
This section provides supplementary information to support the findings presented in the paper. We include descriptions of the game variations and additional results to further validate our conclusions. This information ensures transparency and reproducibility, offering deeper insights into our research methodologies and outcomes.

\subsection{Game Variants/Modifications used in our Evaluation (Section~\ref{sec:experiments}}
\label{app:variants1}
These are the environments, used in our evaluation. We shortly state the changes we made as well as the modifications (\cf Section~\ref{app:modifications}) we used to create these environments. A shorter version can be found in Section~\ref{sec:experiments} and additional environments follow in the next section. 

\begin{enumerate}[leftmargin=20pt,itemsep=1pt,parsep=1pt,topsep=1pt,partopsep=0pt, label=(\roman*)]
    \item \textbf{One Armed \textit{Boxing}.} 
    In \textit{Boxing}, the agent controls the white boxer that has to punch an automated opponent (while avoiding getting punched). In the original game, both boxers can hit their opponents with both arms.
    We introduce \textit{One armed}, where the agent can only use its right arm, forcing the agent to adapt its behavior. 
    This variates the gameplay instead of modifying the color shift, for which human players also might face performance changes.
    \item \textbf{No Barrel \textit{DonkeyKong}.} This version of \textit{DonkeyKong}, we removed the barrels, making it easier to get to Peach. We use the \textit{no\_barrel} modification here.
    \item \textbf{Aligned Cars \textit{Freeway}} This variant aligned the cars so that each car has the same speed. The first 5 cars are driving from left to right in one line. The last 5 from right to left. We used the modification \textit{stop2} for this game variant. 
    \item \textbf{Mono-colored \textit{Freeway}.} In \textit{Freeway}, the agent controls a chicken targeted with crossing a road and is rewarded only after reaching the other side of the road.
    At every contact with an incoming car, the agent is pushed back down, making the probability of crossing (while playing randomly) quite low. \textit{Freeway}'s rewarding system can thus be considered sparse.
    To test for generalization and robustness, we provide multiple color variations, where all cars are colored with the same color using the \textit{color} modification. In \autoref{tab:robustness1} we used \textit{color1}, resulting in black.
    \item \textbf{Stopped Cars Freeway.} While \textit{Aligned Cars Freeway} aligns the cars to move with the same speed, this version stops the cars to make crossing the street even easier. The modification for this variant is \textit{stop3} and can be seen in \autoref{fig:learnable}.
    \item \textbf{No Danger \textit{Frostbite}.} In \textit{Frostbite} there are birds pushing you into the water. To remove this danger and enable skill learning, we used the modification \textit{enemies1}. 
    \item \textbf{Static Ice \textit{Frostbite}.} Another way to loose the game is to jump into the water when missing a floating ice shelf. To make the game a little easier, we use the modification \textit{static60} to fixate the ice shelves.
    \item \textbf{No Danger \textit{Kangaroo}.} In Kangaroo the agent has to learn to deal with monkeys and coconuts that are thrown by the monkeys or are falling from the top floor. In this variant, we removed these elements by using the modifications \textit{disable\_monkeys} and \textit{disable\_coconut}.
    \item \textbf{Swap Level \textit{MsPacman}} MsPacman has multiple levels to play, however, most agents only train on the first level. The difference between the levels primarily changes the layout of the maze. This game variant does now really change the gameplay but lets the agents train or test their ability on another level. We used the \textit{change\_level1} modification for this. 
    \item \textbf{Lazy Enemy \textit{Pong}.} Also explained already in Section~\ref{sec:experiments}, this variant let the enemy stop after hitting the ball and only allows movement again after the player hits the ball with its paddle. The used modification is \textit{lazy\_enemy}.
    \item \textbf{Infinite Oxygen Seaquest} This version has a similar goal as No Enemy Seaquest by removing the necessity to get to the surface to refill oxygen. The player can concentrate on the other tasks. We used the modification \textit{unlimited\_oxygen}.
    \item \textbf{No Shields \textit{SpaceInvaders}.} In SpaceInvaders the user has 3 shields to hide behind. These shields can be removed, using the modification \textit{disable\_shields}. 
    
\end{enumerate}

\subsection{Additional Game Variants}
\label{app:additional_results}
In addition to the variants presented in the paper, we also conducted experiments on the following game variations to show the variety of possible environments, that can be created using HackAtari. This is by no means an exhaustive list and more variants can be created using the modifications, described in \autoref{app:modifications}.

\begin{enumerate}[leftmargin=20pt,itemsep=1pt,parsep=1pt,topsep=1pt,partopsep=0pt, label=(\roman*)]
    \setcounter{enumi}{12}
     \item \textbf{Drunken Boxing.} An alternative to One-Armed Boxing is our Drunken Boxing version. In this variant the player is moved into a random position at each timestep, making the move actions more challenging. For this, we used the \textit{drunken\_boxing} modification. 
    \item \textbf{Gravity Breakout} In Breakout you have to hit the ball before it reaches the bottom of the screen. To make the game more challenging, we added gravity, pulling the ball downwards using an artificial strength. We used the modification \textit{strength1} and \textit{gravity}.
    \item \textbf{Easy FishingDerby} In Fishing Derby the player has to dodge the sharks while catching the fish. In this variant we stopped the sharks from moving and enhanced the amount of fish on the players side. We used the modifications \textit{fish\_mode0} and \textit{shark\_mode0}.
    \item \textbf{No Ghosts MsPacman.} MsPacman has four ghosts trying to catch the player. When catched the player looses one of its lives. In this version of the game, we cage the ghosts to their square in the middle of the maze using \textit{caged\_ghosts}. This version enables the agent to ignore the ghosts completely.
    \item \textbf{No Fuel Riverraid} This versions removes the necessity to collect fuel within the game. The player always have a full tank. The modification is \textit{no\_fuel}. 
    \item \textbf{No Enemy Seaquest} In Seaquest the player has multiple tasks, like saving divers, always have oxygen in the tank and shooting as well as dodging the enemies. In this version we removed the enemies with \textit{disable\_enemies}. 
    \item \textbf{Relocated Shields SpaceInvaders} Instead of removing the shields completely, HackAtari also enables us to relocate them. In this version the shields behave like in the original game but are moves aside slightly. We used \textit{relocate40}.
\end{enumerate}

\begin{figure}[h!]
    \centering
    \includegraphics[width=\linewidth]{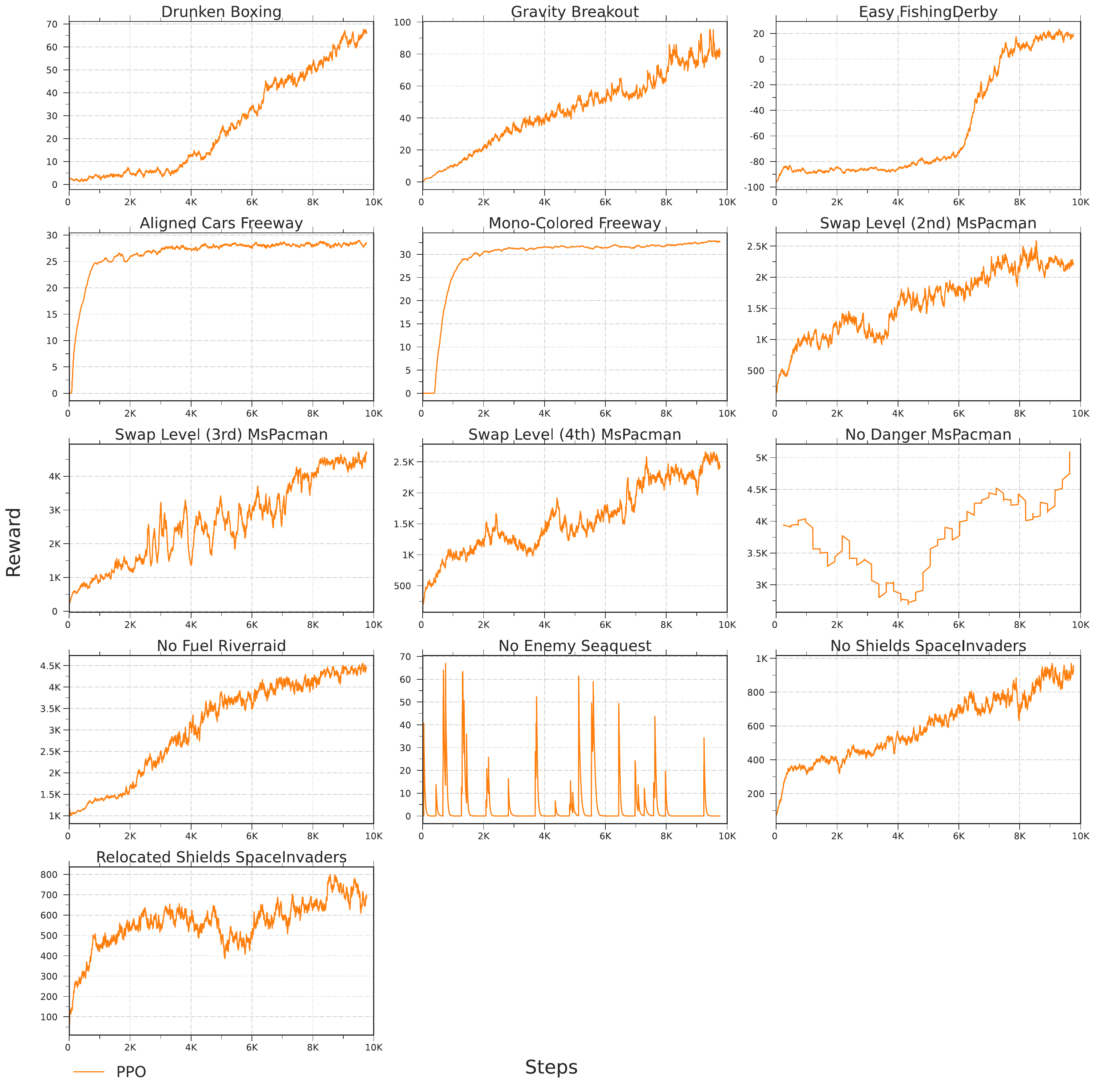}
    \caption{\textbf{Addtional environments RL agents can learn in.} PPO agents are able to progressively improve in these new game variants over 10M training steps.}
    \label{appfig:learnable}
\end{figure}

Similar to \autoref{fig:learnable}, \autoref{appfig:learnable} displays the training process of a PPO agent in these additional environments as well as environments from App.\ref{app:variants1}. It can be seen that PPO agents are able to learn in all of them, except \textit{No Enemy Seaquest}. A reason for the latter can be seen in their sparsity regaring rewards. Without enemies to shoot, the only way to gain rewards is to save 6 divers. This is similar Kangaroo in \autoref{tab:robustness1}.

\subsection{Extended Experimental Setup}
In \autoref{tab:robustness1}, we compared agents trained in the original environment against their performance in some of HackAtari's new variants. The performance of a similar agent, trained in the game variation instead of the original environment, is also added to show that higher scores are reachable. The performances were measured over a span of 30 episodes. Next to ending an episode naturally by winning or failing the game, we also add a maximum number of 100000 frames after which the episode is truncated as well as a maximum reward of 50000 after which the game is also truncated. As error we displayed the standard deviation over all episodes.

Fig.~\ref{fig:learnable} to \ref{fig:skill_learning} display training runs over 10M frames/steps. The exact hyperparameters can be found in App.~\ref{app:hyperparameter}. As rewards we used the internal ones given by the Atari games and for Fig.~\ref{fig:rfchanged} added our own reward function as additional axis. To mitigate noise and fluctuations, we use exponential moving average (EMA) smoothing over the mean of all seeds. We use an effective window size of \( 50 \), resulting in a smoothing factor \( \alpha = 2 / (1 + 50) \approx 0.039 \) used in the following formula:
\begin{align}
     \mathit{EMA}_t = (1-\alpha) \cdot \mathit{EMA}_{t-1} + \alpha \cdot y_t
     \,\text.
\end{align}
To manage irregular training intervals due to rewards are not always being reported in the same timestep, we ignore missing values when computing the average, relying on the EMA smoothing to provide a continuous curve. For the error bands we used the standard deviation of your data within a rolling window over all seeds.

\subsection{Evaluating Human Performance on HackAtari Environments}
\label{app:human_eval}
All experiments conducted in this study that involve evaluating human performance strictly adhere to ethical guidelines. 
We place the utmost importance on the anonymity and confidentiality of all participants. Data collected during the study is anonymized, ensuring that no personal information can be linked back to the participants.

The goal of this small study was to investigate, how humans react to unknown changes in familiar Atari games and the impact of these changes on game outcomes. The data collected, including game interaction data and results, will be used solely for scientific purposes. We are committed to conducting this research with the highest ethical standards, respecting the privacy and rights of all participants. 

Participants were asked to make themselves familiar with a selected set of games by playing them until they felt comfortable in them. Games were chosen randomly per participant. Then we evaluated their performance in each of the original environments for 5 minutes before switching to the HackAtari variant of that specific game. We also set a cap in reward to 50000 in games like Kangaroo or DonkeyKong where our game variant removed the natural cause of death and truncated the game if a player reaches this reward. Note that participants have not seen or played the variant before and had no prior training time in it, compared to the original environment. 

The results in \autoref{tab:robustness1} show the mean over all participants. All participants were informed about the study before taking part in it. This includes potential risks, usage of information recorded (\ie the performance), the tasks within the study. The were also informed that withdrawal was possible at all times, resulting in the removal of all personal information (\ie the performance scores recorded until withdrawal). All participants had to give their consent before taking part. There was no compensation for participation. 

The study was taken under the guidelines of the institutional ethics committee. 

\section{Reproducing Training and Train new Models}
Hackatari provides a robust environment for training reinforcement learning agents on a wide range of game modifications. By keeping it modular and near the original gymnasium environment, we emphasize leveraging frameworks like Stable Baselines and CleanRL, to develop and fine-tune agents to excel in customized game scenarios. This section will walk you through the steps to set up training processes and adapt to various game dynamics using Hackatari. We emphasize the importance of reproducibility, ensuring that experiments can be reliably replicated and validated. For this, we will also provide any needed information of our specific training, starting with the hyperparameters. 

\subsection{Hyperparameter Configuration for Training}
\label{app:hyperparameter}
In this section, we provide a detailed overview of the hyperparameters employed during the training and optimization of our models. Hyperparameters play a pivotal role in determining the performance and generalization ability of machine learning models. In our experiments, we followed the parameter set by \citet{huang2022cleanrl} for both our C51 and PPO agents. We do not provide any grid search or hyperparameter optimization.

\begin{table}[h]
\caption{\textbf{Hyperparameter Configuration for Experimental Settings (PPO).} This table provides a comprehensive overview of the essential hyperparameters utilised in our experimental section.}

\label{tab:hyperparams1}
\centering
\begin{tabular}{llll}
\toprule
\textbf{Hyperparameter} & \textbf{Value} & \textbf{Hyperparameter} & \textbf{Value}  \\
\midrule
batch size              & 1024           & Clipping Coef.          & 0.1               \\
$\gamma$                & 0.99           & KL target               & None                \\
minibatch size          & 256            & GAE lambda              & 0.95               \\
seeds                   & 42,73,91       & input representation    & 4x84x84            \\
total timesteps         & 10M            & gym version             & 0.28.1            \\
learning rate           & 0.00025        & pytorch version         & 1.12.1            \\
more information        & \multicolumn{3}{l}{\small \url{https://docs.cleanrl.dev/rl-algorithms/ppo/}}\\

\bottomrule
\end{tabular}
\end{table}

\begin{table}[h]
\caption{\textbf{Hyperparameter Configuration for Experimental Settings (C51).} This table provides a comprehensive overview of the essential hyperparameters utilized in our experimental section. }
\label{tab:hyperparams2}
\centering
\begin{tabular}{llll}
\toprule
\textbf{Hyperparameter} & \textbf{Value} & \textbf{Hyperparameter} & \textbf{Value}  \\
\midrule
batch size              & 32             & optimizer               & Adam               \\
buffer size             & 100k           & loss                    & cross-entropy loss \\
$\gamma$                & 0.99           & input representation    & 4x84x84            \\
seeds                   & 42,73,91       & gym version             & 0.28.1            \\
total time steps        & 10M            & pytorch version         & 1.12.1            \\
learning rate           & 0.00025        &                         &                    \\
more information        & \multicolumn{3}{l}{\small \url{https://docs.cleanrl.dev/rl-algorithms/c51/}}\\

\bottomrule
\end{tabular}
\end{table}

\subsection{Hardware Specification and Computational Costs}

\begin{table*}[tbh!]
    \centering
     \caption{Hardware configuration for our experimental section. (NVIDIA DGX-2 Working station) }
    \label{tab:hardware}
    \begin{tabular}{lr} \toprule
        \textbf{Hardware} & \textbf{Description}  \\ \midrule
        CPU & Intel(R) Xeon(R) Platinum 8174 CPU @ 3.10GHz \\
        GPU & 16 $\times$ NVIDIA® Tesla V100 \\
        Memory & 1.5 TB 2,133 MHz DDR4 RDIMM  \\
        Operating System & Ubuntu 20.04 LTS \\
        \bottomrule
    \end{tabular}
   
\end{table*}

The experiments were run on a setup, described in \autoref{tab:hardware}, using the NVIDIA GPU Cloud (NGC) docker container for pytorch\footnote{\url{https://catalog.ngc.nvidia.com/orgs/nvidia/containers/pytorch}, accessed 2024-05-22}. As stated before, all needed data is openly available. 

For the PPO agents, the main load of training these networks was done on the CPU. While training the process did 400 to 600 steps per second (SPS), resulting in a overall runtime of 5-7h per agent per seed. Since we were running the experiments on a shared server, this of course was highly depending on the general load of the machine. For this paper we trained round about 70-100 agents (including failed runs). 
C51 agents were trained using the GPUs to accelerate. When training an agent on one of the V100 GPUs, training took about 10-12h. Since C51 were only used as an alternative to PPO for some experiments, we only trained about 15 agents. Overall this results in about 750 hours of training time.

\subsection{Training and Evaluating Agents using HackAtari}
Node that HackAtari alone is more a wrapper for gymnasium which can then be included in training environments like Stable-baselines or CleanRL. 

To reproduce our training, one needs to replace the gymnasium in CleanRL with HackAtari. We provide\footnote{\url{https://github.com/BluemlJ/oc_cleanrl}} a fork already doing this for C51 and PPO. (See \textit{ppo\_atari.py} and \textit{c51\_atari.py}). The training and evaluation process follows CleanRL, using its codebase for tracking and logging results and models. 

Examples on how our agents are trained and evaluated, can be found in the scripts folder, together with a ReadMe about how to train your own agents and a small installation guide in form of a bash script. For this it is neccessary to first download the HackAtari repository\footnote{\url{https://github.com/k4ntz/HackAtari}} and add it to the subfolder.  

The codebase for training will be released together with the HackAtari repository.

\section{Replacing the Reward Function}
\label{app:changingtheRF}

In this section, we demonstrate the process of replacing the reward function in HackAtari.

By customizing the reward function, you can tailor the reinforcement learning environment to better suit your research objectives and explore new dimensions of agent behavior, e.g., like being a pacifistic agent. We'll cover the steps needed to implement your own reward structure, ensuring you can effectively modify the learning incentives within the game.

\paragraph{Example: Skiing.}
In the Atari game "Skiing," the reward function is based on the player's score, which is determined by the time taken to complete the course and the penalties incurred for missing gates. Here is a more detailed breakdown of how the reward is typically calculated:
\begin{itemize}
    \item \textbf{Time-Based Scoring} The primary goal in Skiing is to navigate through a series of gates as quickly as possible. The faster the player completes the course, the higher the score.
        The game keeps track of the time taken to finish the course. A lower time translates into a better score.
    \item \textbf{Gate Penalties}
        Players must pass through gates correctly to avoid penalties.
        Each missed gate results in a penalty, which usually adds extra time to the player's overall time. These penalties decrease the score since the total time increases.
\end{itemize}
    
However, the result is giving to the agent as a combined score after each episode making it hard to learn that game. The result a non-functional agent, not able to play the game at all. To enable agents to learn the game however, we propose to change the reward function over the training process, starting with simple skills like skiing downwards before adding poles to it.

To use an own reward function, one can provide the path to a valid python file including a function called \textit{reward\_function}. 

\begin{minted}[mathescape,
               linenos,
               numbersep=5pt,
               gobble=0,
               frame=lines,
               framesep=2mm]{python}
LAST_SCORE=32 # save the RAM value from the last turn

def reward_function(self) -> float:
    global LAST_SCORE 
    score = self.get_ram()[107] #nr. of gates successfully passed
    if score != LAST_SCORE:
        reward = 100
    else:
        reward = 0
    LAST_SCORE = score
    return reward
\end{minted}

The example above shows a simple reward function only rewarding passing successfully through gates and sets the reward to 0 for every other turn. The number of successfully passed gates is saved as a decreasing number from 32 to 0 in the RAM at coordinate $107$. The complexity of the new reward function depends only on the user and can also be created using LLMs or game objects. 

\paragraph{Generating a reward function with a LLM in Seaquest.} The following example of a reward function, for the game of Seaquest, was created using ChatGPT 3.5 and the results can be seen in \autoref{fig:rfchanged}. For the input to the LLM, we used game objects, object properties and a short description of the game.
The object-centric environment context is given by the classes provided by the OCAtari framework \citep{delfosse2023ocatari}, i.e., the parent game object class\footnote{ \url{https://github.com/k4ntz/OC_Atari/blob/master/ocatari/ram/game_objects.py}} and the game-specific objects \footnote{e.g., \url{https://github.com/k4ntz/OC_Atari/blob/master/ocatari/ram/pong.py}}. The game objects related to the score were omitted. The game description is taking from the Atari description\footnote{\url{https://www.retrogames.cz/play_221-Atari2600.php}}, the gymnasium website\footnote{\url{https://gymnasium.farama.org/environments/atari/seaquest/}} and some shortening from us. For the game of Seaquest the description is the following

\begin{minted}[mathescape,
               linenos,
               numbersep=5pt,
               gobble=0,
               frame=lines,
               framesep=2mm,
               breaksymbolleft={},
               tabsize=2,breaklines]{text}
You a sub (Player) able to move in all directions and fire torpedoes.
The goal is to retrieve as many divers as you can, while dodging and blasting enemy subs and killer sharks.
The game begins with one sub and three waiting on the horizon. Each time you increase your score by 10,000 points, an extra sub will be delivered to your base. 
Your sub will explode if it collides with anything except your divers.The sub has a limited amount of oxygen that decreases at a constant rate during the game. When the oxygen tank is almost empty, you need to surface and if you don't do it in time, your sub will blow up and you'll lose one diver. 
Each time you're forced to surface, with less than six divers, you lose one diver as well.
\end{minted}

The object-centric reward function created by the LLM for Seaquest:

\begin{minted}[mathescape,
               linenos,
               numbersep=5pt,
               gobble=0,
               frame=lines,
               framesep=2mm,
               linenos,tabsize=2,breaklines]{python}
from ocatari.ram.seaquest import *

def check_collision(obj1, obj2):
    """
    Check if two GameObjects collide based on their bounding boxes.
    """
    return (obj1.x < obj2.x + obj2.w and
            obj1.x + obj1.w > obj2.x and
            obj1.y < obj2.y + obj2.h and
            obj1.y + obj1.h > obj2.y)

def update_game_state(objects):
    """
    Update positions of all game objects and check for collisions.
    """
    collisions = []
    for obj in objects:
        # Update position based on velocity
        obj.xy = (obj.x + obj.dx, obj.y + obj.dy)
        
        # Check for collisions with other objects
        for other in objects:
            if obj != other and check_collision(obj, other):
                collisions.append((obj, other))
    return collisions

def manage_oxygen_and_lives(player, oxygen_bar, lives):
    """
    Decrease oxygen levels and manage lives based on oxygen and collisions.
    """
    # Decrease oxygen
    oxygen_bar.value -= 1
    if oxygen_bar.value <= 0:
        player.lives -= 1
        oxygen_bar.value = 100  # Reset oxygen after surfacing or losing a life
    
    # Check if lives are depleted
    if lives.value <= 0:
        print("Game Over")

def update_score_and_divers(player, divers_collected, score):
    """
    Update score based on collected divers and manage divers.
    """
    for diver in divers_collected:
        if check_collision(player, diver):
            score.value += 1000  # Increment score for each diver collected
            divers_collected.remove(diver)  # Remove diver from the game

def fire_torpedo(player, torpedoes):
    """
    Create a new torpedo at the player's location and add it to the torpedoes list.
    """
    new_torpedo = PlayerMissile()
    new_torpedo.xy = player.xy
    torpedoes.append(new_torpedo)


def reward_function(self) -> float:
    game_objects = self.game_objects()
    reward = 0.0
    
    # Define categories for easy identification
    player = None
    divers = []
    enemies = []
    player_missiles = []
    enemy_missiles = []
    oxygen_bar = None
    
    # Classify objects
    for obj in game_objects:
        if isinstance(obj, Player):
            player = obj
        elif isinstance(obj, Diver):
            divers.append(obj)
        elif isinstance(obj, Shark) or isinstance(obj, Submarine):
            enemies.append(obj)
        elif isinstance(obj, PlayerMissile):
            player_missiles.append(obj)
        elif isinstance(obj, EnemyMissile):
            enemy_missiles.append(obj)
        elif isinstance(obj, OxygenBar):
            oxygen_bar = obj
    
    # Check for collisions and manage interactions
    if player:
        for diver in divers:
            if check_collision(player, diver):
                reward += 0.1  # Scaled down reward for collecting a diver
                divers.remove(diver)  # Assume diver is collected and removed from the game
        
        for enemy in enemies:
            if check_collision(player, enemy):
                reward -= 0.1  # Scaled down penalty for colliding with an enemy
        
        for missile in enemy_missiles:
            if check_collision(player, missile):
                reward -= 0.05  # Scaled down penalty for getting hit by an enemy missile
        
        # Reward for hitting enemies with missiles
        for missile in player_missiles:
            for enemy in enemies:
                if check_collision(missile, enemy):
                    reward += 0.05  # Scaled down reward for destroying an enemy
                    enemies.remove(enemy)  # Assume enemy is destroyed and removed from the game
                    player_missiles.remove(missile)  # Remove missile after hitting

    # Manage oxygen levels
    if oxygen_bar and oxygen_bar.value <= 20:
        reward -= 0.05  # Scaled down penalty for low oxygen levels
    
    # Encourage surfacing if oxygen is too low
    if oxygen_bar and oxygen_bar.value <= 10:
        reward -= 0.1  # Scaled down higher penalty for critically low oxygen

    return reward
\end{minted}

\clearpage

\section{Modifications}
\label{app:modifications}

This chapter delves into the extensive range of modifications HackAtari offers, providing users with a versatile toolkit to customize and enhance their reinforcement learning experiments. From changing the game dynamics and creating new challenges to integrating novel algorithms, we explore the myriad ways HackAtari can be adapted to meet diverse research needs.

\subsection{Game-specific Modifications}

\subsubsection{BankHeist \ details}
\IfFileExists{environment_description/BankHeist.tex}{
\begin{minipage}{0.60\textwidth}
You are a bank robber and (naturally) want to rob as many banks as possible. You control your getaway car and must navigate maze-like cities. The police chases you and will appear whenever you rob a bank. You may destroy police cars by dropping sticks of dynamite. You can fill up your gas tank by entering a new city.At the beginning of the game you have four lives. Lives are lost if you run out of gas, are caught by the police,or run over the dynamite you have previously dropped.
\end{minipage}
}{}
\IfFileExists{environment_images/BankHeist.png}{
\begin{minipage}{0.38\textwidth}
\raggedleft
\includegraphics[scale=0.25]{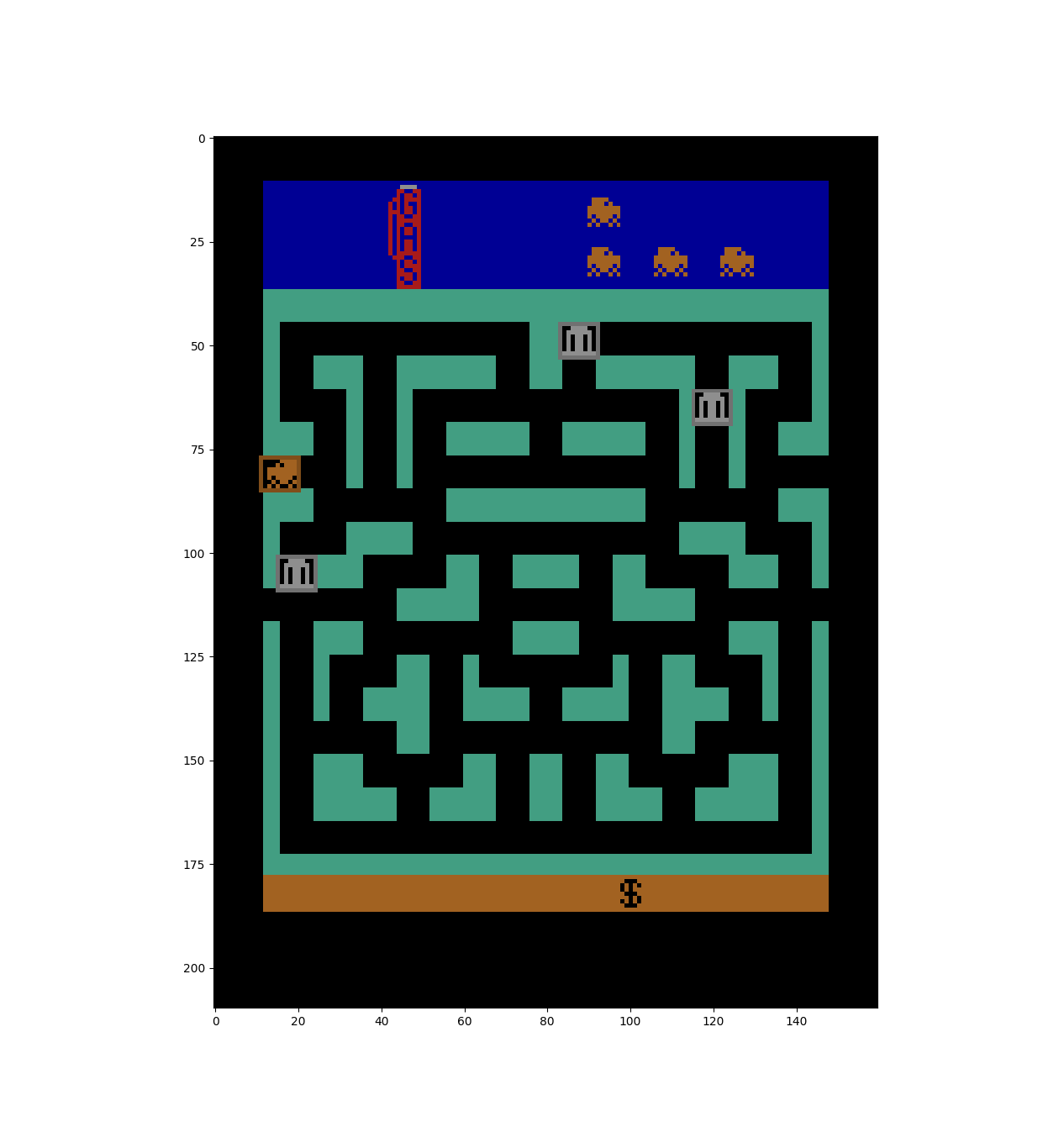}
\end{minipage}{}
}{}

\foreach \n in {BankHeist}{\input{reports/\n.tex}}

\subsubsection{BattleZone \ details}
\IfFileExists{environment_description/BattleZone.tex}{
\begin{minipage}{0.60\textwidth}
You control a tank and must destroy enemy vehicles. This game is played in a first-person perspective and creates a 3D illusion. A radar screen shows enemies around you. You start with 5 lives and gain up to 2 extra lives if you reach a sufficient score.
\end{minipage}
}{}
\IfFileExists{environment_images/BattleZone.png}{
\begin{minipage}{0.38\textwidth}
\raggedleft
\includegraphics[scale=0.25]{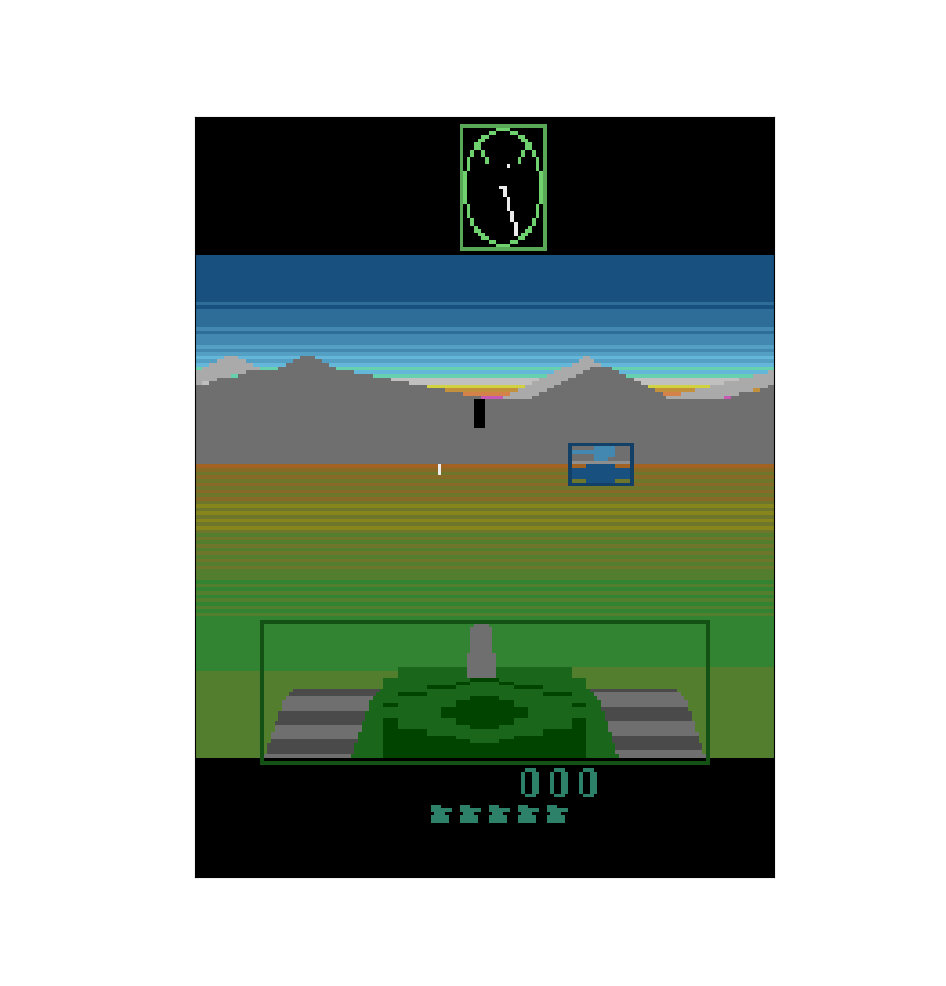}
\end{minipage}{}
}{}

\subsubsection{Boxing \ details}
\IfFileExists{environment_description/Boxing.tex}{
\begin{minipage}{0.60\textwidth}
You fight an opponent in a boxing ring. You score points for hitting the opponent. If you score 100 points, your opponent is knocked out.
\end{minipage}
}{}
\IfFileExists{environment_images/Boxing.png}{
\begin{minipage}{0.38\textwidth}
\raggedleft
\includegraphics[scale=0.25]{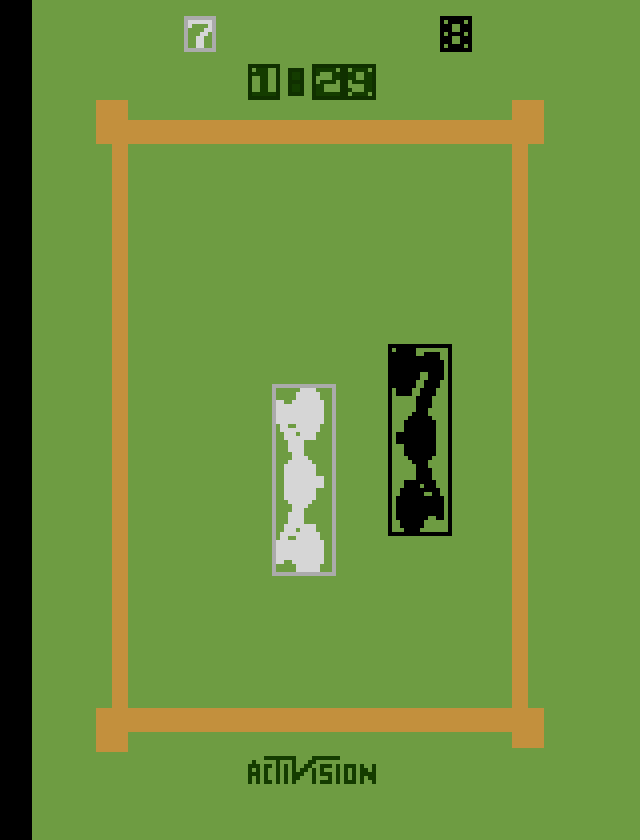}
\end{minipage}{}
}{}

\begin{table}[h]
    \centering
    \begin{tabular}{p{3cm} | m{2cm} | p{8cm}} \toprule
        \textbf{Modification} & \textbf{Parameter} & \textbf{Effect} \\ \midrule
        gravity$X$ & $X \in (1,5)$ & Add a permanent downwards movement with strength $X$. \\
        one\_armed &&  Disables the "hitting motion" with the right arm permanently \\
        drunken\_boxing && Applies random movements to the players input. \\
        color\_p$X$ &$X \in (0,4)$&  Changes the color of the player to [Black, White, Red, Blue, Green] by choosing a value 0-4 \\
        color\_e$X$ &$X \in (0,4)$&  Changes the color of the enemy to [Black, White, Red, Blue, Green] by choosing a value 0-4 \\
        switch\_p && Switches the position of player and enemy \\
    \bottomrule
    \end{tabular}
\end{table}

\subsubsection{Breakout \ details}
\IfFileExists{environment_description/Breakout.tex}{
\begin{minipage}{0.60\textwidth}
Another famous Atari game. The dynamics are similar to pong: You move a paddle and hit the ball in a brick wall at the top of the screen. Your goal is to destroy the brick wall. You can try to break through the wall and let the ball wreak havoc on the other side, all on its own! You have five lives.
\end{minipage}
}{}
\IfFileExists{environment_images/Breakout.png}{
\begin{minipage}{0.38\textwidth}
\raggedleft
\includegraphics[scale=0.25]{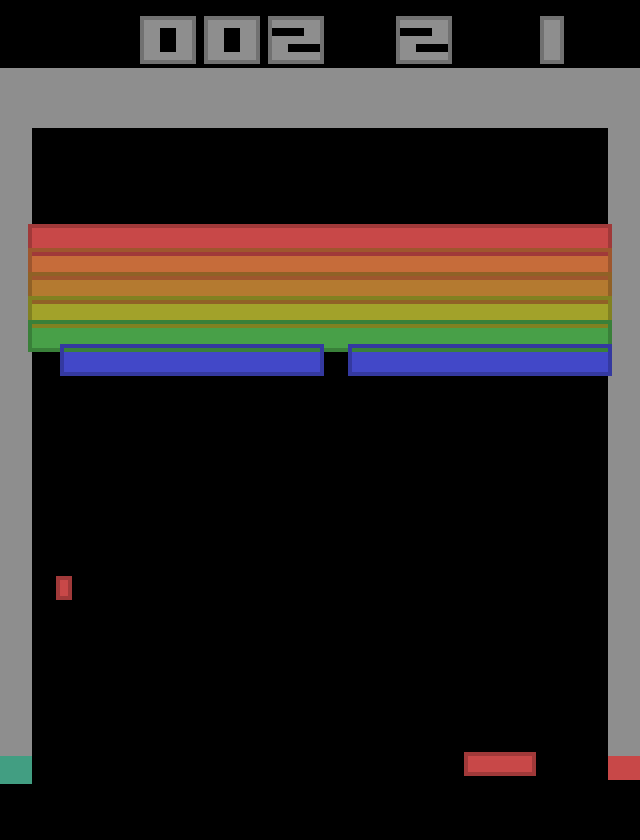}
\end{minipage}{}
}{}

\begin{table}[h]
    \centering
    \begin{tabular}{p{3cm} | m{2cm} | p{8cm}} \toprule
        \textbf{Modification} & \textbf{Parameter} & \textbf{Effect} \\ \midrule
        strength$X$ & $X$ & Set strength of drift to $X$. \\
        drift$X$ &$X \in \{r,l\}$&  Set drift direction to left or right.\\
        gravity && Pull the ball down by changing the corresponding ram positions \\
        inverse\_gravity && Pushes the ball up by changing the corresponding ram positions \\
        color\_p$X$ &$X \in (0,4)$&  Changes the color of the player to 
        [Black, White, Red, Blue, Green] by choosing a value 0-4 \\
        color\_b$X$ &$X \in (0,4)$&  Changes the color of all blocks to [Black, White, Red, Blue, Green] by choosing a value 0-4 \\
        color\_r$XY$ &$X,Y \in (0,4)$ &  Changes the color of row $X$ to color $Y$ [Black, White, Red, Blue, Green] by choosing a value 0-4 for both row and color \\
        
    \bottomrule
    \end{tabular}
\end{table}

\subsubsection{Carnival \ details}
\IfFileExists{environment_description/Carnival.tex}{
\begin{minipage}{0.60\textwidth}
This is a “shoot ‘em up” game. Targets move horizontally across the screen and you must shoot them. You are in control of a gun that can be moved horizontally. The supply of ammunition is limited and chickens may steal some bullets from you if you don’t hit them in time.
\end{minipage}
}{}
\IfFileExists{environment_images/Carnival.png}{
\begin{minipage}{0.38\textwidth}
\raggedleft
\includegraphics[scale=0.25]{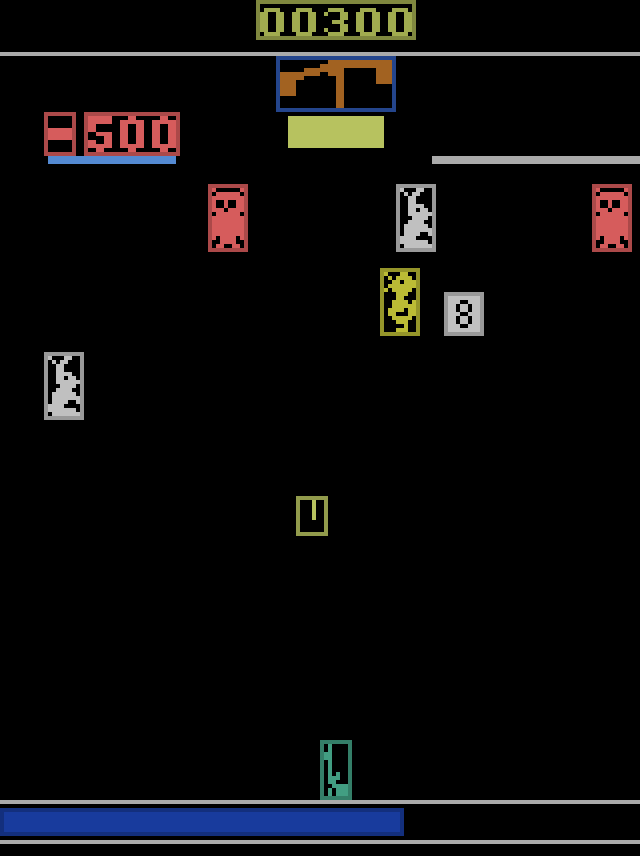}
\end{minipage}{}
}{}

\begin{table}[h]
    \centering
    \begin{tabular}{p{3cm} | m{2cm} | p{8cm}} \toprule
        \textbf{Modification} & \textbf{Parameter} & \textbf{Effect} \\ \midrule
        no\_flying\_ducks && Ducks in the last row disappear instead of turning into flying ducks. \\
        unlimited\_ammo && Ammunition doesn't decrease. \\
        fast\_missiles$X$ &$X \in (1,3)$& The projectiles fired from the players are faster. Uses the values 1-3 to determine how by how much to speed the missile up.\\
        
    \bottomrule
    \end{tabular}
\end{table}

\subsubsection{ChopperCommand \ details}
\IfFileExists{environment_description/ChopperCommand.tex}{
\begin{minipage}{0.60\textwidth}
You control a helicopter and must protect truck convoys. To that end, you need to shoot down enemy aircraft.A mini-map is displayed at the bottom of the screen.
\end{minipage}
}{}
\IfFileExists{environment_images/ChopperCommand.png}{
\begin{minipage}{0.38\textwidth}
\raggedleft
\includegraphics[scale=0.25]{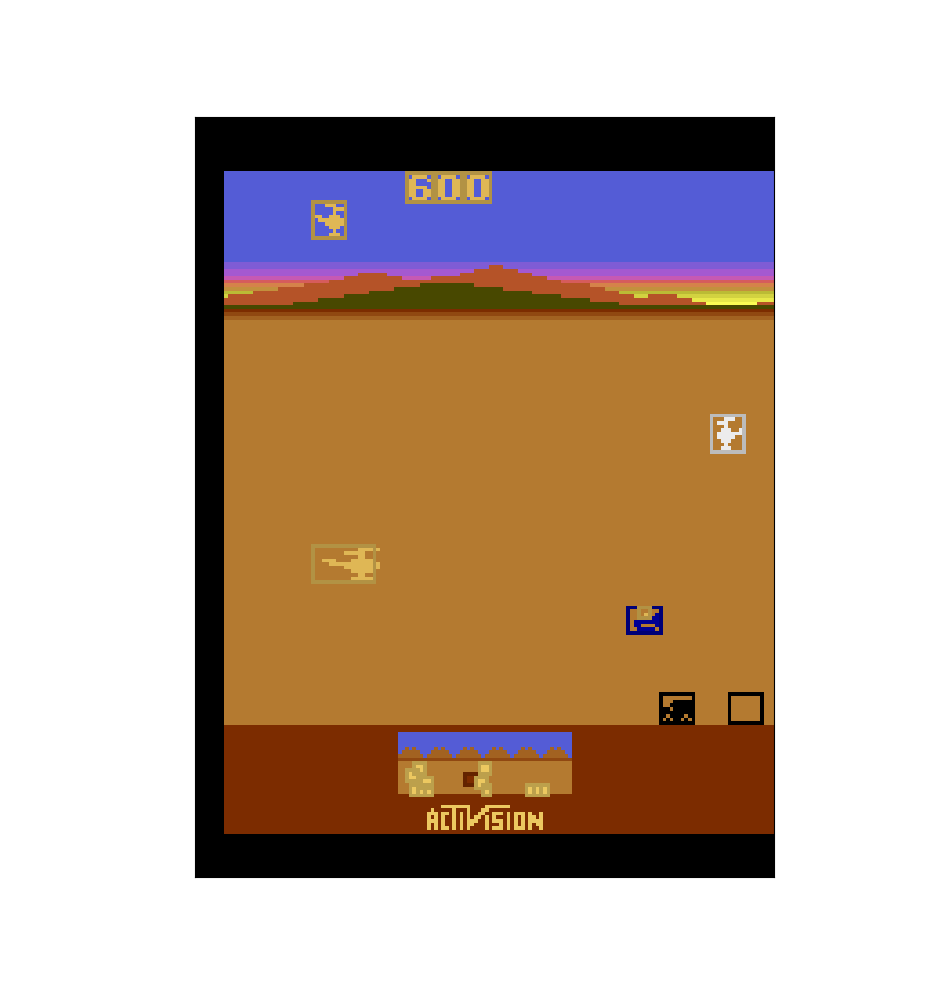}
\end{minipage}{}
}{}

\begin{table}[h]
    \centering
    \begin{tabular}{p{3cm} | m{2cm} | p{8cm}} \toprule
        \textbf{Modification} & \textbf{Parameter} & \textbf{Effect} \\ \midrule
        delay\_shots && Puts time delay between shots \\
        no\_enemies && Removes all Enemies from the game\\
        no\_radar && Removes the radar content\\
        invis\_player && Makes the player invisible \\
        color$X$ &$X \in (0,4)$& Changes the color of background to [Black, White, Red, Blue, Green] by choosing a value 0-4. This also affects the enemies colors \\
        \bottomrule
    \end{tabular}
\end{table}

\subsubsection{DonkeyKong \ details}
\IfFileExists{environment_description/DonkeyKong.tex}{
\begin{minipage}{0.60\textwidth}
You play as Mario trying to save your girlfriend who has been kidnapped by Donkey Kong. Remove rivets and jump over fireballs, with a score that starts high and counts down throughout the game.
\end{minipage}
}{}
\IfFileExists{environment_images/DonkeyKong.png}{
\begin{minipage}{0.38\textwidth}
\raggedleft
\includegraphics[scale=0.25]{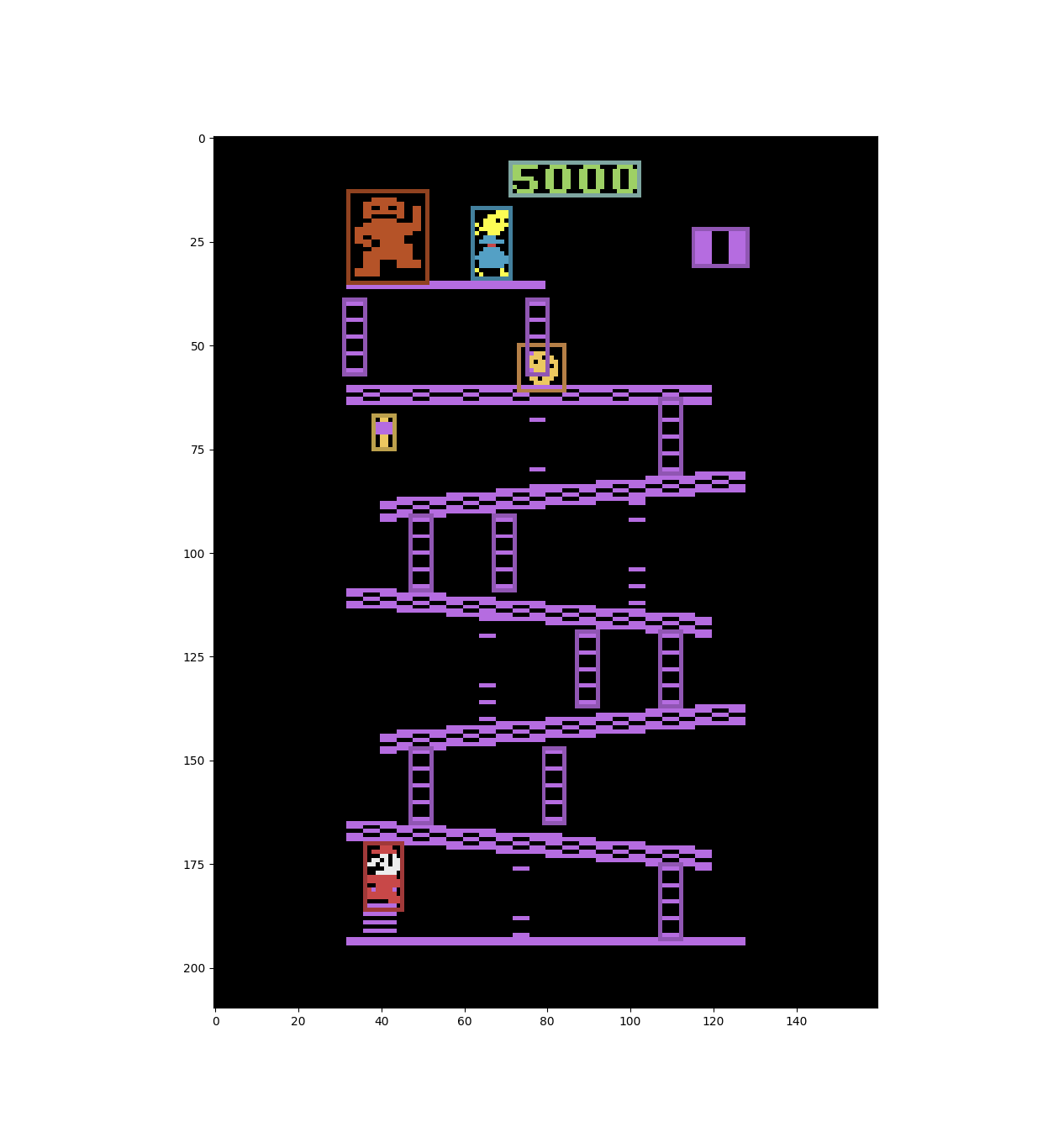}
\end{minipage}{}
}{}

\begin{table}[h]
    \centering
    \begin{tabular}{p{3cm} | p{10cm}} \toprule
        \textbf{Modification} & \textbf{Effect} \\ \midrule
       no\_barrel & Remove barrels from the game \\
       random\_start & Set the start position to a random pre-defined start position \\
        \bottomrule
    \end{tabular}
\end{table}

\subsubsection{FishingDerby \ details}
\IfFileExists{environment_description/FishingDerby.tex}{
\begin{minipage}{0.60\textwidth}
Your objective is to catch more sunfish than your opponent.
\end{minipage}
}{}
\IfFileExists{environment_images/FishingDerby.png}{
\begin{minipage}{0.38\textwidth}
\raggedleft
\includegraphics[scale=0.25]{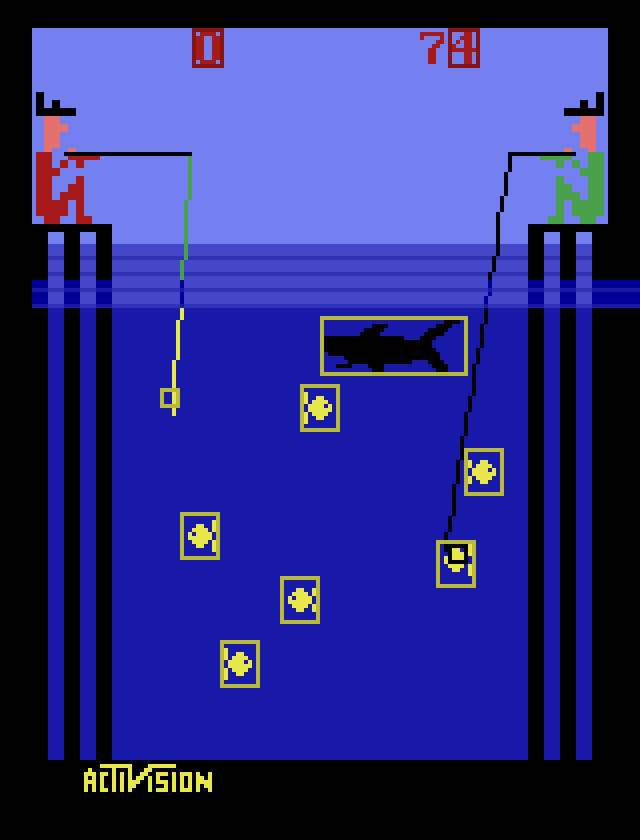}
\end{minipage}{}
}{}

\begin{table}[h]
    \centering
    \begin{tabular}{p{3cm} | m{2cm} | p{8cm}} \toprule
        \textbf{Modification} & \textbf{Parameter} & \textbf{Effect} \\ \midrule
        fish\_mode$X$ &$X \in \{0,1,3\}$ & Allows for alterations in the behavior of the shark as specified \\
        shark\_mode$X$ &$X \in (0,3)$ & Allows for alterations in the behavior of the fish as specified \\
        \bottomrule
    \end{tabular}
\end{table}

\subsubsection{Freeway \ details}
\IfFileExists{environment_description/Freeway.tex}{
\begin{minipage}{0.60\textwidth}
Your objective is to guide your chicken across lane after lane of busy rush hour traffic. You receive a point for every chicken that makes it to the top of the screen after crossing all the lanes of traffic.
\end{minipage}
}{}
\IfFileExists{environment_images/Freeway.png}{
\begin{minipage}{0.38\textwidth}
\raggedleft
\includegraphics[scale=0.25]{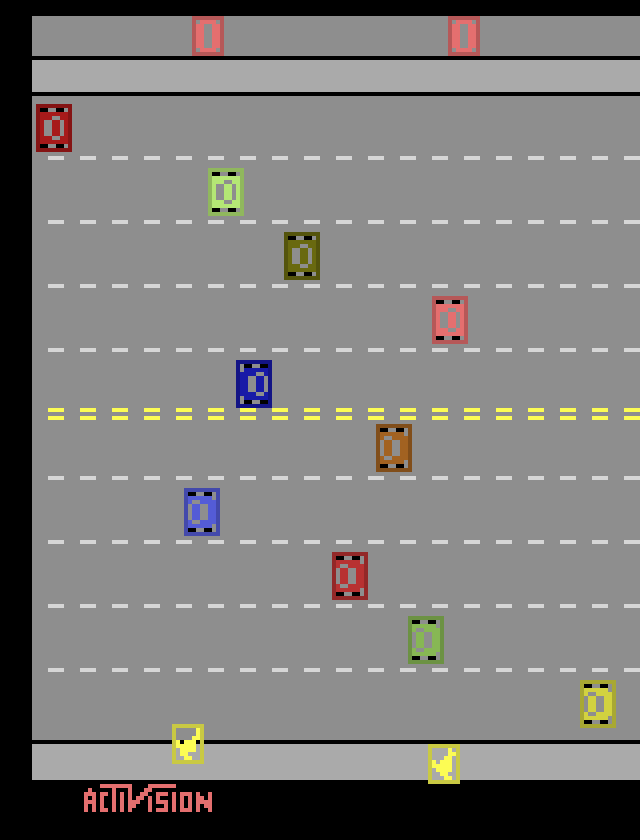}
\end{minipage}{}
}{}

\begin{table}[h]
    \centering
    \begin{tabular}{p{3cm} | m{2cm} | p{8cm}} \toprule
        \textbf{Modification} & \textbf{Parameter} & \textbf{Effect} \\ \midrule
        stop$X$ & $X \in (1,3)$ & Manipulate the move pattern of the car or let them stop.\\
        color$X$ & $X \in (0,8)$ & Set color of cars to one of 9 pre-defined colors \\
        \bottomrule
    \end{tabular}
\end{table}

\subsubsection{Frostbite \ details}
\IfFileExists{environment_description/Frostbite.tex}{
\begin{minipage}{0.60\textwidth}
In Frostbite, the player controls “Frostbite Bailey” who hops back and forth across across an Arctic river, changing the color of the ice blocks from white to blue. Each time he does so, a block is added to his igloo.
\end{minipage}
}{}
\IfFileExists{environment_images/Frostbite.png}{
\begin{minipage}{0.38\textwidth}
\raggedleft
\includegraphics[scale=0.25]{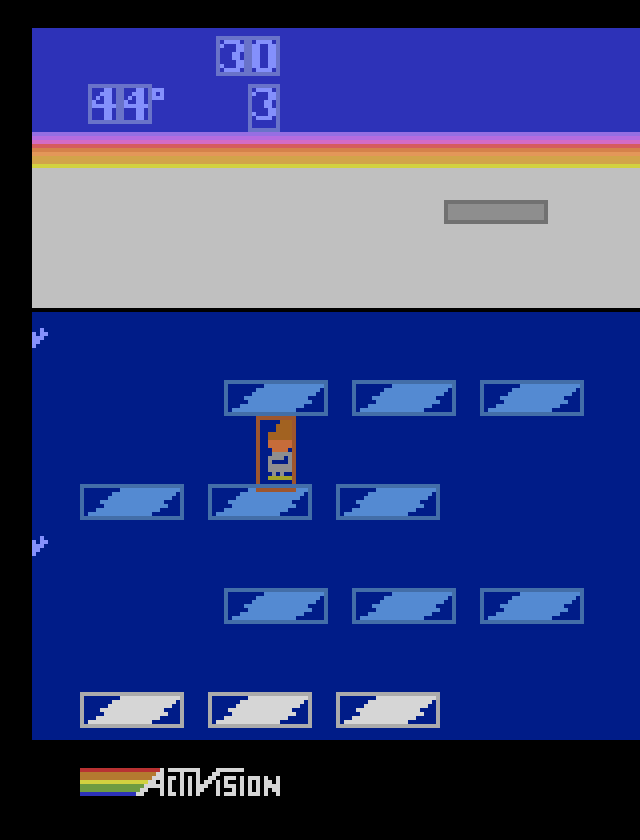}
\end{minipage}{}
}{}

\begin{table}[h]
    \centering
    \begin{tabular}{p{3cm} | m{2cm} | p{8cm}} \toprule
        \textbf{Modification} & \textbf{Parameter} & \textbf{Effect} \\ \midrule
        color$X$ &$X \in (0,3)$ & Adjusts the colors of the ice floes bases on the specified values.\\
        line$X$ &$X \in (1,5)$& Specify row/line for color change \\
        ui\_color$X$ &$X \in (0,3)$ &Adjust color of the UI \\
        enemies$X$ &$X \in (0,3)$& Adjusts the memory based on the specified number of enemies selected by the user. \\
        floes$XXX$ &$X \in (0,160)$&  Adjusts the memory based on the specified new position of the ice floes.\\
        static$XXX$ &$X \in (0,160)$&  Adjusts the memory based on the specified new position of the ice floes and hold it in place for the game.\\
       
        \bottomrule
    \end{tabular}
\end{table}

\subsubsection{Kangaroo \ details}
\IfFileExists{environment_description/Kangaroo.tex}{
\begin{minipage}{0.60\textwidth}
The object of the game is to score as many points as you can while controlling Mother Kangaroo to rescue her precious baby. You start the game with three lives.During this rescue mission, Mother Kangaroo encounters many obstacles. You need to help her climb ladders, pick bonus fruit, and throw punches at monkeys.
\end{minipage}
}{}
\IfFileExists{environment_images/Kangaroo.png}{
\begin{minipage}{0.38\textwidth}
\raggedleft
\includegraphics[scale=0.25]{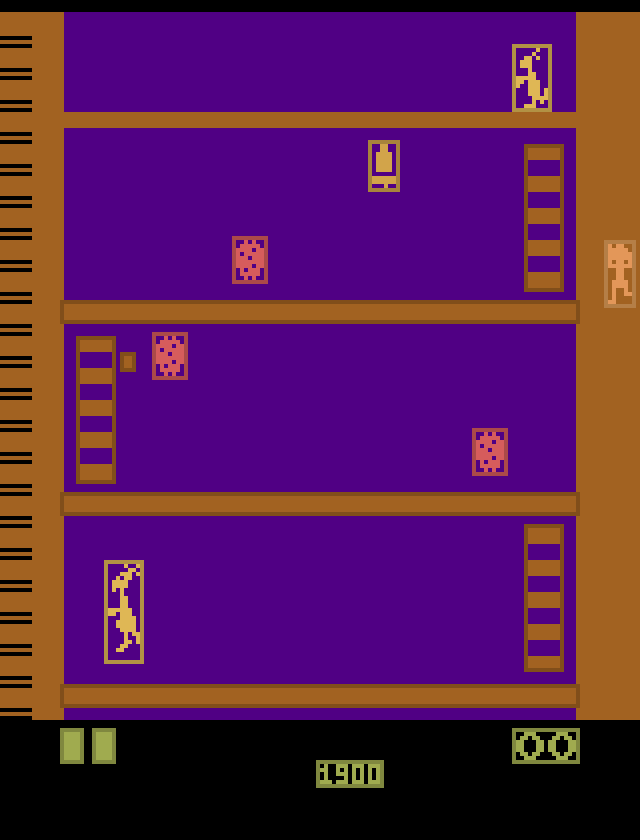}
\end{minipage}{}
}{}

\begin{table}[h]
    \centering
    \begin{tabular}{p{3cm} | m{2cm} | p{8cm}} \toprule
        \textbf{Modification} & \textbf{Parameter} & \textbf{Effect} \\ \midrule
        disable\_monkeys && Disables the monkeys in the game\\
        disable\_coconut && Disables the coconuts in the game \\
        random\_init && Randomize the floor on which the player starts.\\
        set\_floor$X$ &$X \in (0,2)$& Set the floor on which the player starts.\\
        change\_level$X$ &$X \in (0,2)$&  Changes the level according to the argument number 0-2. \textit{change\_level}, selects random level. \\
        \bottomrule
    \end{tabular}
\end{table}

\subsubsection{MontezumaRevenge \ details}
\IfFileExists{environment_description/MontezumaRevenge.tex}{
\begin{minipage}{0.60\textwidth}
Your goal is to acquire Montezuma’s treasure by making your way through a maze of chambers within the emperor’s fortress. You must avoid deadly creatures while collecting valuables and tools which can help you escape with the treasure.
\end{minipage}
}{}
\IfFileExists{environment_images/MontezumaRevenge.png}{
\begin{minipage}{0.38\textwidth}
\raggedleft
\includegraphics[scale=0.25]{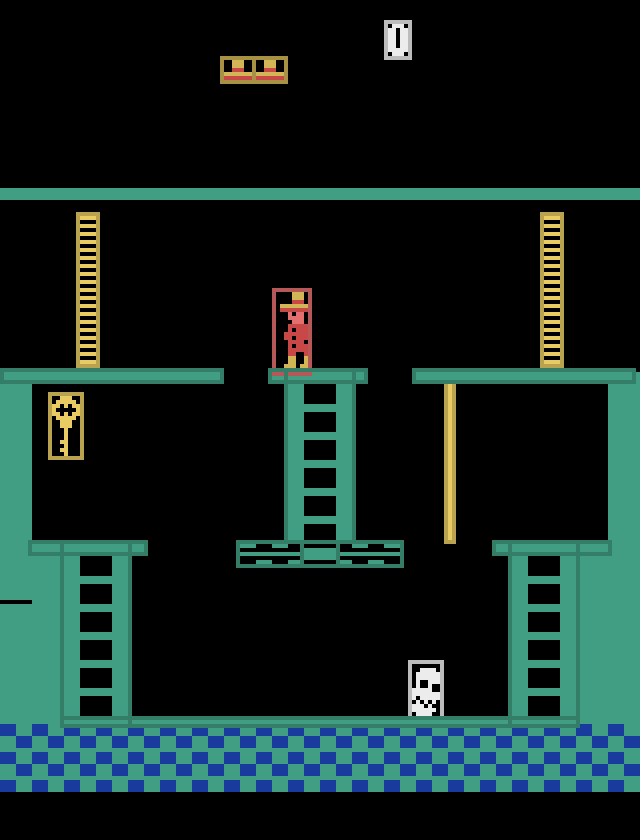}
\end{minipage}{}
}{}

\begin{table}[h]
    \centering
    \begin{tabular}{p{3cm} | m{2cm} | p{8cm}} \toprule
        \textbf{Modification} & \textbf{Parameter} & \textbf{Effect} \\ \midrule
        random\_position\_start && Randomize the start position within the room \\
        level$X$ &$X \in (0,9)$& Changes the level to a more difficult version. Level 0, 1, 2 are different versions, afterwards level determines map layout. \\
        randomize\_items && Randomize which item is found in which room. \\
        full\_inventory && Adds all items to inventory. \\
       
        \bottomrule
    \end{tabular}
\end{table}

\subsubsection{MsPacman \ details}
\IfFileExists{environment_description/MsPacman.tex}{
\begin{minipage}{0.60\textwidth}
Your goal is to collect all of the pellets on the screen while avoiding the ghosts.
\end{minipage}
}{}
\IfFileExists{environment_images/MsPacman.png}{
\begin{minipage}{0.38\textwidth}
\raggedleft
\includegraphics[scale=0.25]{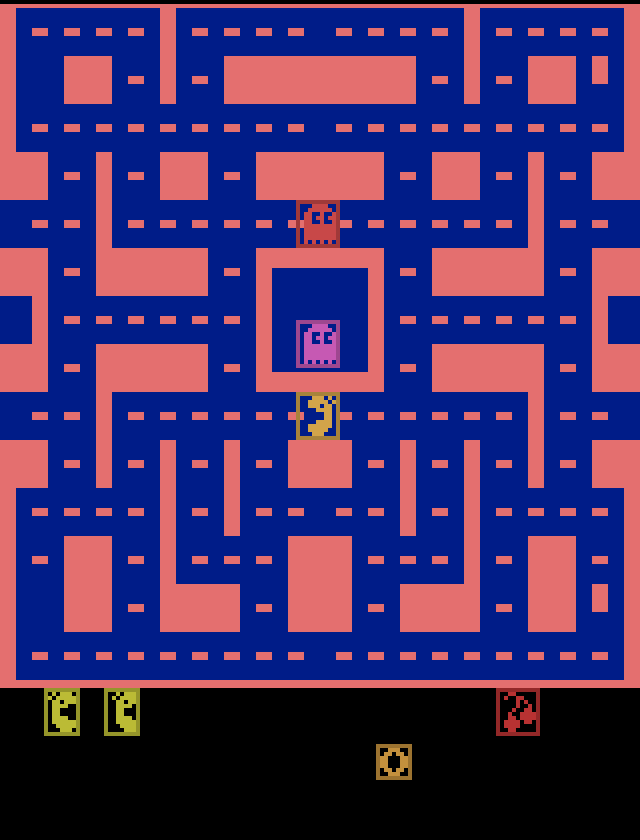}
\end{minipage}{}
}{}

\begin{table}[h]
    \centering
    \begin{tabular}{p{3cm} | m{2cm} | p{8cm}} \toprule
        \textbf{Modification} & \textbf{Parameter} & \textbf{Effect} \\ \midrule
        caged\_ghosts && Fix the position of
    the ghost inside the square in the middle of the screen. \\
        disable\_orange && Fix the position of the orange ghost only. \\
        disable\_red && Fix the position of the red ghost only. \\
        disable\_cyan && Fix the position of the cyan ghost only. \\
        disable\_pink && Fix the position of the pink ghost only. \\
        power$X$ &$X \in (0,4)$& Switching the specified number of power pills with normal edible tokens.\\
        edible\_ghosts && All ghost will be made edible the entire game \\
        inverted && All ghost will be edible the entire game until the player eats a power pill. After eating a power pill the ghost will return to "normal" for a certain amount of time \\
        change\_level$X$ &$X \in (0,3)$& Changes the level according to the argument number 0-3. \\
        
        \bottomrule
    \end{tabular}
\end{table}

\subsubsection{Pong \ details}
\IfFileExists{environment_description/Pong.tex}{
\begin{minipage}{0.60\textwidth}
You control the right paddle, you compete against the left paddle controlled by the computer. You each try to keep deflecting the ball away from your goal and into your opponent’s goal.
\end{minipage}
}{}
\IfFileExists{environment_images/Pong.png}{
\begin{minipage}{0.38\textwidth}
\raggedleft
\includegraphics[scale=0.25]{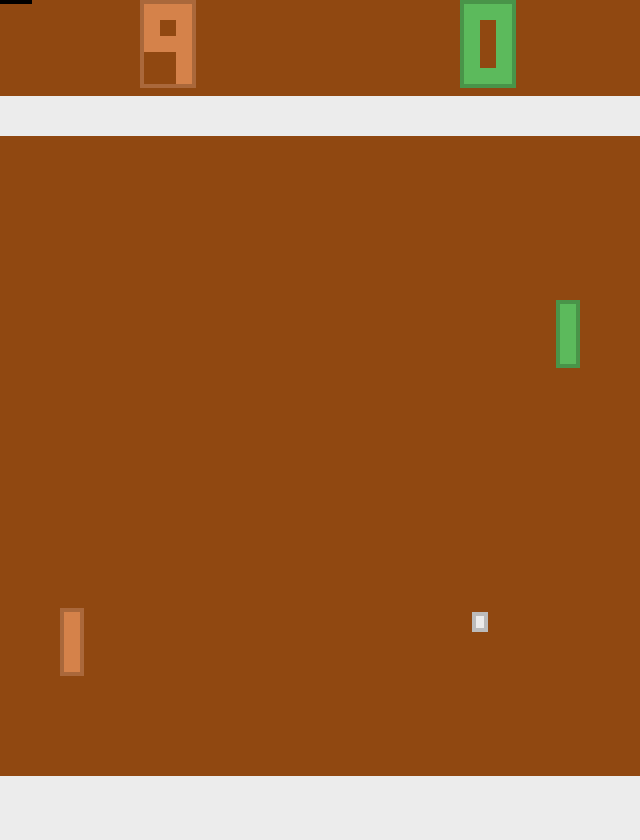}
\end{minipage}{}
}{}

\begin{table}[h]
    \centering
    \begin{tabular}{p{3cm} | m{2cm} | p{8cm}} \toprule
        \textbf{Modification} & \textbf{Parameter} & \textbf{Effect} \\ \midrule
        lazy\_enemy && Enemy does not move after returning the shot until player hits ball.\\
        up\_drift$X$ &$X \in (1,5)$&  Makes the ball drift upwards by changing the corresponding ram positions \\
        down\_drift$X$ &$X \in (1,5)$&  Makes the ball drift down by changing the corresponding ram positions\\
        left\_drift$X$ &$X \in (1,5)$&  Makes the ball drift to the left by changing the corresponding ram positions\\
        right\_drift$X$ &$X \in (1,5)$&  Makes the ball drift to the right by changing the corresponding ram positions\\
        \bottomrule
    \end{tabular}
\end{table}

\subsubsection{RiverRaid \ details}
\IfFileExists{environment_description/RiverRaid.tex}{
\begin{minipage}{0.60\textwidth}
\begin{table}[h]
    \centering
    \begin{tabular}{p{3cm} | p{10cm}} \toprule
        \textbf{Modification} & \textbf{Effect} \\ \midrule
        no\_fuel & Removes the fuel deposits from the game. \\
        \bottomrule
    \end{tabular}
\end{table}
\end{minipage}
}{}
\IfFileExists{environment_images/RiverRaid.png}{
\begin{minipage}{0.38\textwidth}
\raggedleft
\includegraphics[scale=0.25]{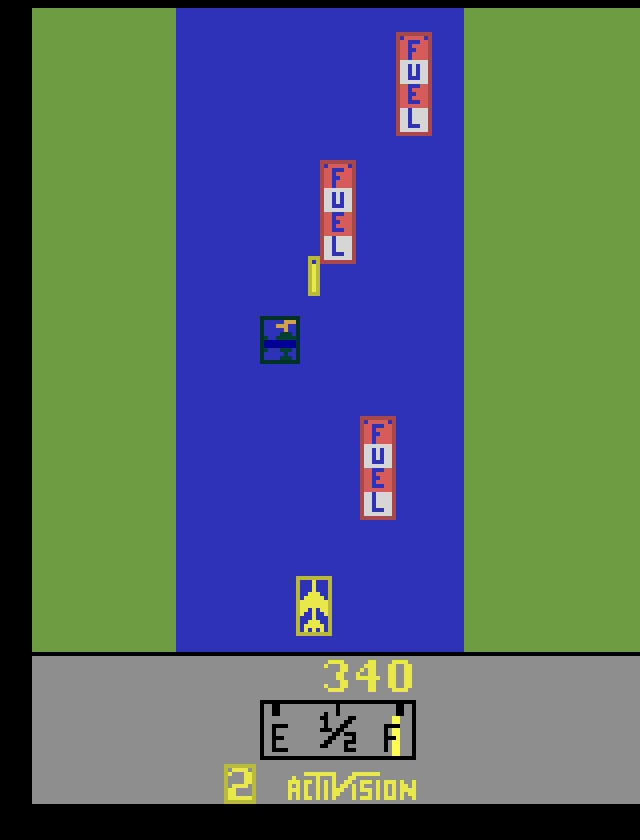}
\end{minipage}{}
}{}

\subsubsection{Seaquest \ details}
\IfFileExists{environment_description/Seaquest.tex}{
\begin{minipage}{0.60\textwidth}
You control a sub able to move in all directions and fire torpedoes. The goal is to retrieve as many divers as you can, while dodging and blasting enemy subs and killer sharks; points will be awarded accordingly. The game begins with one sub and three waiting on the horizon. Each time you increase your score by 10,000 points, an extra sub will be delivered to your base. You can only have six reserve subs on the screen at one time.Your sub will explode if it collides with anything except your own divers.The sub has a limited amount of oxygen that decreases at a constant rate during the game. When the oxygen tank is almost empty, you need to surface and if you don’t do it in time, your sub will blow up and you’ll lose one diver. Each time you’re forced to surface, with less than six divers, you lose one diver as well.
\end{minipage}
}{}
\IfFileExists{environment_images/Seaquest.png}{
\begin{minipage}{0.38\textwidth}
\raggedleft
\includegraphics[scale=0.25]{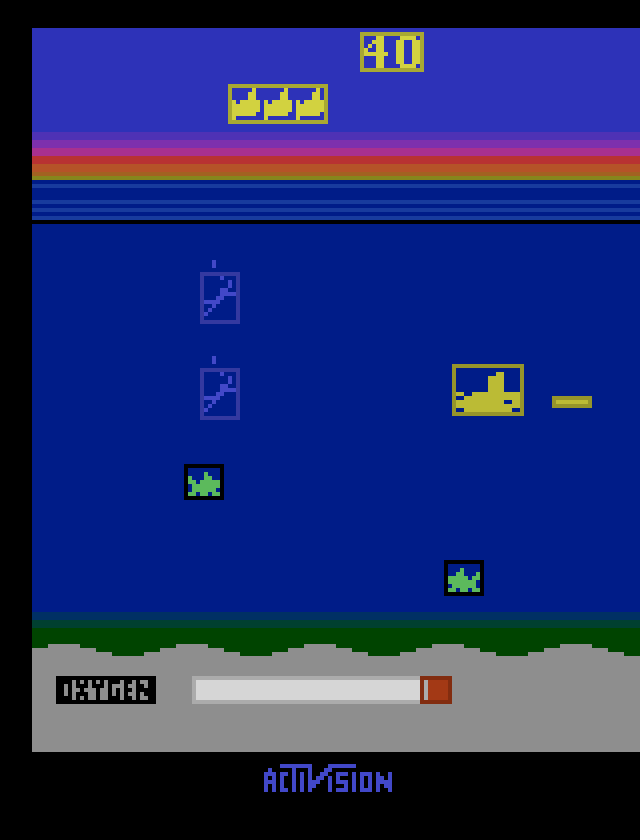}
\end{minipage}{}
}{}

\foreach \n in {Seaquest}{\input{reports/\n.tex}}

\subsubsection{Skiing \ details}
\IfFileExists{environment_description/Skiing.tex}{
\begin{minipage}{0.60\textwidth}
You control a skier who can move sideways. The goal is to run through all gates (between the poles) in the fastest time. You are penalized five seconds for each gate you miss. If you hit a gate or a tree, your skier will jump back up and keep going.
\end{minipage}
}{}
\IfFileExists{environment_images/Skiing.png}{
\begin{minipage}{0.38\textwidth}
\raggedleft
\includegraphics[scale=0.25]{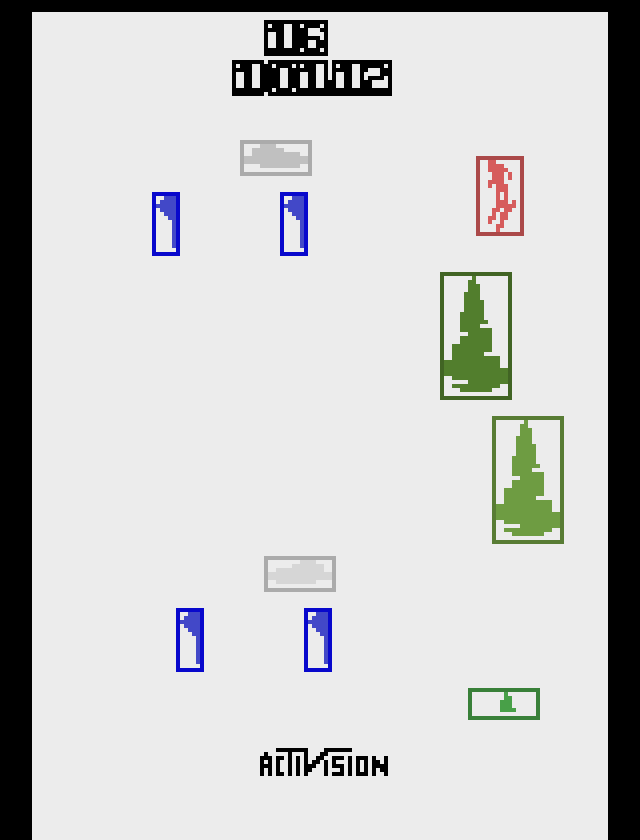}
\end{minipage}{}
}{}

\begin{table}[h]
    \centering
    \begin{tabular}{p{3cm} | p{10cm}} \toprule
        \textbf{Modification} & \textbf{Effect} \\ \midrule
       invert\_flags & Switches the flag color from blue to red\\
        \bottomrule
    \end{tabular}
\end{table}

\subsubsection{SpaceInvaders \ details}
\IfFileExists{environment_description/SpaceInvaders.tex}{
\begin{minipage}{0.60\textwidth}
Your objective is to destroy the space invaders by shooting your laser cannon at them before they reach the Earth. The game ends when all your lives are lost after taking enemy fire, or when they reach the earth.
\end{minipage}
}{}
\IfFileExists{environment_images/SpaceInvaders.png}{
\begin{minipage}{0.38\textwidth}
\raggedleft
\includegraphics[scale=0.25]{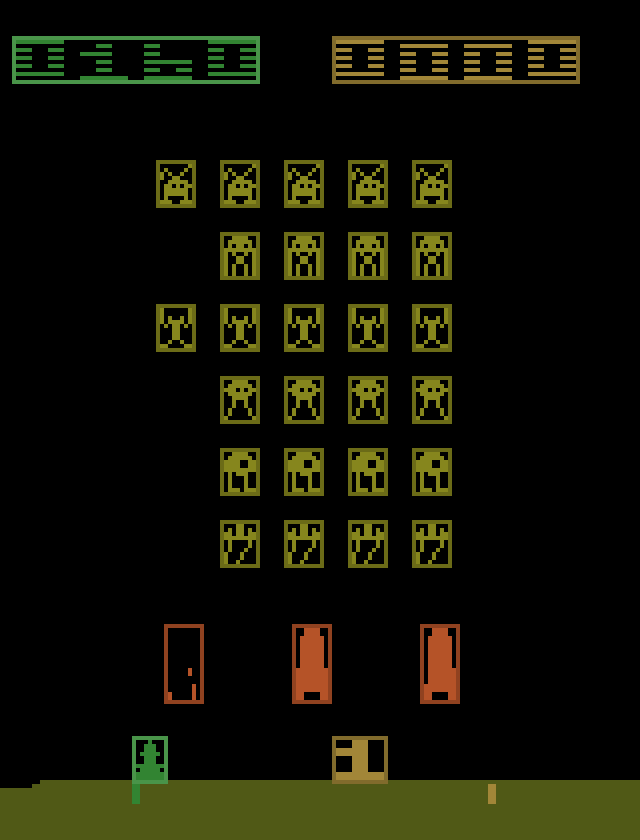}
\end{minipage}{}
}{}

\foreach \n in {SpaceInvaders}{\input{reports/\n.tex}}

\subsubsection{Tennis \ details}
\IfFileExists{environment_description/Tennis.tex}{
\begin{minipage}{0.60\textwidth}
You control the orange player playing against a computer-controlled blue player. The game follows the rules of tennis. The first player to win at least 6 games with a margin of at least two games wins the match. If the score is tied at 6-6, the first player to go 2 games up wins the match.
\end{minipage}
}{}
\IfFileExists{environment_images/Tennis.png}{
\begin{minipage}{0.38\textwidth}
\raggedleft
\includegraphics[scale=0.25]{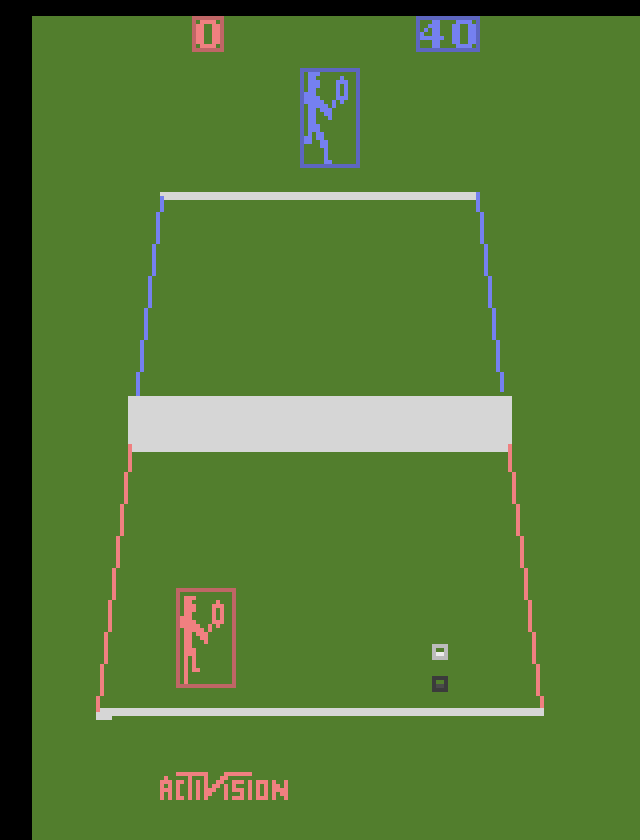}
\end{minipage}{}
}{}

\begin{table}[h]
    \centering
    \begin{tabular}{p{3cm} | p{10cm}} \toprule
        \textbf{Modification} & \textbf{Effect} \\ \midrule
        wind & Sets the ball in the up and right direction by 3 pixles every single ram step to simulate the effect of wind \\
        upper\_pitches & Changes the ram so that it is always the upper persons turn to pitch. \\
        lower\_pitches & Changes the ram so that it is always the lower persons turn to pitch. \\
        upper\_player & Changes the ram so that the player is always in the upper field \\
        lower\_player & Changes the ram so that the player is always in the lower field \\
        
        \bottomrule
    \end{tabular}
\end{table}

\end{document}